\documentclass [10pt,twocolumn]{IEEEtran}
\usepackage{algorithm,algorithmic,amsbsy,amsmath,amssymb,epsfig,bbm,mathrsfs,multirow,amsthm}
\usepackage[T1]{fontenc}
\usepackage{fixltx2e}
\usepackage{algorithm}
\usepackage{array,multirow,graphicx}
\usepackage{color}
\usepackage{cite}
\usepackage{float}
\usepackage{soul}
\usepackage{makecell}
\usepackage[x11names,dvipsnames,table]{xcolor} 
\usepackage{colortbl}
\usepackage{multirow}
\usepackage{subcaption} 
\usepackage{lipsum}
\usepackage{booktabs}
\usepackage[french, german, vietnamese, english]{babel}
\definecolor{mygray}{gray}{0.6}
\definecolor{myblue}{rgb}{0.8,0.85,1} 
\usepackage{array}
\newcolumntype{L}[1]{>{\raggedright\let\newline\\\arraybackslash\hspace{0pt}}m{#1}}
\newcolumntype{C}[1]{>{\centering\let\newline\\\arraybackslash\hspace{0pt}}m{#1}}
\newcolumntype{R}[1]{>{\raggedleft\let\newline\\\arraybackslash\hspace{0pt}}m{#1}}

\usepackage{array, tabularx, boldline}
\usepackage{eurosym}
\usepackage{amstext} 
\DeclareRobustCommand{\officialeuro}{%
  \ifmmode\expandafter\text\fi
  {\fontencoding{U}\fontfamily{eurosym}\selectfont e}}

\usepackage{mathtools}
\usepackage{array, tabularx, boldline}
\usepackage{graphicx}
\usepackage{cellspace}
\usepackage[bookmarks=false]{hyperref}
\begin{document}
\title{\huge Diffusion Models for Future Networks and Communications: A Comprehensive Survey}
\author{Nguyen Cong Luong, Nguyen Duc Hai, \textit{Student Member, IEEE}, Duc Van Le, \textit{Senior Member, IEEE}, \\Huy T. Nguyen, \textit{Member, IEEE}, Thai-Hoc Vu, Thien Huynh-The, \textit{Senior Member, IEEE}, \\Ruichen Zhang, Nguyen Duc Duy Anh, Dusit Niyato, \textit{Fellow, IEEE}, Marco Di Renzo, \textit{Fellow, IEEE}, \\Dong In Kim, \textit{Fellow, IEEE}, and Quoc-Viet Pham, \textit{Senior Member, IEEE}

\thanks{Nguyen Cong Luong, Nguyen Duc Hai and Nguyen Duc Duy Anh are with the Phenikaa School of Computing, Phenikaa University, Hanoi 12116, Vietnam. E-mails: luong.nguyencong@phenikaa-uni.edu.vn,  21010560@st.phenikaa-uni.edu.vn, and 21011488@st.phenikaa-uni.edu.vn.}

\thanks{Duc Van Le is with the School of Electrical Engineering and Telecommunications, University of New South Wales, NSW 2052, Australia. E-mail: duc.le1@unsw.edu.au.}

\thanks{Huy T. Nguyen is with Smart and Autonomous Systems Research Group, Faculty of Information Technology, School of Technology, Van Lang University, Ho Chi Minh City, 70000, Vietnam. Email: huy.nt@vlu.edu.vn.}

\thanks{Thai-Hoc Vu is with the Institute of Information Technology, Digital Transformation, Thu Dau Mot University, Binh Duong 820000, Vietnam (e-mail: vuthaihoc1995@gmail.com).}

\thanks{Thien Huynh-The is with the Department of Computer and Communications Engineering, Ho Chi Minh City University of Education and Technology, Ho Chi Minh City 71307, Vietnam. E-mail: thienht@hcmute.edu.vn.}

\thanks{Ruichen Zhang and Dusit Niyato are with the College of Computing and Data Science, Nanyang
Technological University, Singapore 639798. E-mails: ruichen.zhang@ntu.edu.sg and dniyato@ntu.edu.sg.}

\thanks{M. Di Renzo is with Université Paris-Saclay, CNRS, CentraleSupélec, Laboratoire des Signaux et Systèmes, 3 Rue Joliot-Curie, 91192 Gif-sur-Yvette, France. (marco.di-renzo@universite-paris-saclay.fr), and with King’s College London, Centre for Telecommunications Research – Department of Engineering, WC2R 2LS London, United Kingdom (marco.di renzo@kcl.ac.uk).
}

\thanks{Dong In Kim is with the Department of Electrical and Computer Engineering, Sungkyunkwan University, Suwon 16419, South Korea. E-mail: dongin@skku.edu.}

\thanks{Quoc-Viet Pham is with the School of Computer Science and Statistics, Trinity College Dublin, Dublin 2, D02 PN40, Ireland. Email: viet.pham@tcd.ie.}
}




\maketitle
\begin{abstract}
The rise of Generative AI (GenAI) in recent years has catalyzed transformative advances in wireless communications and networks. Among the members of the GenAI family, Diffusion Models (DMs) have risen to prominence as a powerful option, capable of handling complex, high-dimensional data distribution, as well as consistent, noise-robust performance. In this survey, we aim to provide a comprehensive overview of the theoretical foundations and practical applications of DMs across future communication systems. We first provide an extensive tutorial of DMs and demonstrate how they can be applied to enhance optimizers, reinforcement learning and incentive mechanisms, which are popular approaches for problems in wireless networks. Then, we review and discuss the DM-based methods proposed for emerging issues in future networks and communications, including channel modeling and estimation, signal detection and data reconstruction, integrated sensing and communication, resource management in edge computing networks, semantic communications and other notable issues. We conclude the survey with highlighting
technical limitations of DMs and their applications, as well as discussing future research directions.

\end{abstract}

\section{Introduction} 
\label{sec:introduction}


In recent years, the emergence of Generative AI (GenAI) has offered a potential paradigm that transcends the capabilities of traditional AI \cite{10679152}. GenAI focuses on the high-level generation of new data across various scales \cite{10506539}. GenAI and its wide range of applications have left a significant impact on important aspects of modern society, including business conduct \cite{korzynski2023generative}, natural science research \cite{reddy2025towards} and social services \cite{baldassarre2023social}. 

GenAI is principally an ensemble of numerous different model families, each with its own unique advantages, drawbacks and contributions to the advancement of AI. The most prominent GenAI models are Variational Autoencoders (VAEs) \cite{kingma2013auto}, Generative Adversarial Networks (GANs) \cite{goodfellow2020gan} and Diffusion Models (DMs) \cite{sohl2015deep}. 
Among those families, DMs have emerged as the most prominent one, overtaking VAEs and GANs in a variety of domains, most notably computer vision \cite{croitoru2023diffusion}, natural language processing \cite{zou2023survey} and and multi-modal modeling \cite{jiang2024survey}. The popularity of DM-based generation applications, such as Stable Diffusion \cite{stable_diff} which sees more than 10 million users daily, is a concrete proof of the superiority of DMs over the remaining members of the GenAI collection. DMs become prevalent as a result of their exceptional capability of modeling and synthesizing complex data distributions \cite{cao2024diffsurvey}, as well as stable training processes and effectiveness when incorporating conditional guidance \cite{rombach2022high}. Furthermore, DMs have shown the adaptability to various types of data \cite{cao2024diffsurvey}, while Transformer-based and autoregressive models are specifically designed for sequential data. 

DMs have recently been proposed to solve various issues in future wireless communication systems. First, DMs are capable of learning complex data distributions, making them a powerful tool for modeling the stochasticity of wireless channels and channel estimation \cite{lee2024generating, sengupta2023generative}. Second, DMs' noise perturbation and removal approach for data recovery works particularly well in low SNR conditions. Thus, they offer an ideal alternative for current signal detection and data reconstruction methods \cite{he2024massive, zhao2024signal}. Third, DMs are also highly impervious to noise and are able to provide high-quality synthetic data to overcome training limitations. As a result, they are highly effective for solving issues in Integrated Sensing and Communication (ISAC) systems, such as signal detection and target detection \cite{wen2024generative, xu2024conditional,Nuria2024}. Fourth, DMs have been known as an efficient alternative for policy and data synthesis in Deep Reinforcement Learning (DRL) \cite{10032267}, guaranteeing better exploitation and exploration in DRL frameworks. Thus, they have recently been the prominent choice for resource management problems in edge computing (EC) networks \cite{cheng2024dependency, xu2024accelerating}. Additionally, DMs have emerged as a promising solution for semantic communication issues \cite{qin2024ai, getu2025semantic} due to their intrinsic denoising capabilities and high-level data generation \cite{wang2024temporal, 10531073}. Specifically, DMs allow high-quality semantic data reconstruction and enhance the fidelity and semantic integrity of data transmitted through channels. Other key research issues such as wireless security \cite{lv2024safeguarding, su2024hybrid} have also been addressed with the integration of DMs. 

There have been some interesting surveys regarding GenAI and DMs. However, they do not provide a comprehensive surveillance on the applications of DMs for networks and communications. Particularly, the survey in \cite{celik2024dawn} provides an exhaustive review regarding GenAI in 6G wireless systems, outlining the integrations of general GenAI models in pioneering areas of 6G network research such as semantic communications, ISAC and THz communications. Meanwhile, the survey in \cite{chen2024generative} particularly focuses on the utilization of general GenAI techniques for human digital twin (HDT) in IoT-healthcare applications,  including personalized health monitoring and diagnosis, prescription, and rehabilitation. The work in \cite{emami2025diffusion} investigates the integration of DMs to enhance reinforcement learning (RL) decision-making and digital modeling for UAV communications. A holistic system architecture of mobile AI generative content (AIGC)-driven HDT is proposed in \cite{chen2024revolution}, alongside rigorous analysis of key design requirements and challenges in applying AIGC-driven HDT for personalized healthcare. The work in \cite{liu2024integration} discusses the integration of federated learning and GenAI models and highlights the threats to centralized federated GenAI models regarding data privacy, integrity and availability. The authors further demonstrate the unique benefits of blockchain in decentralized federated GenAI models in addressing these issues. Regarding energy harvesting - a promising solution to resource-constrained IoT systems, the authors in \cite{xie2024generative} discuss energy harvesting technologies based on renewable natural source and RF energy source with their issues that can be addressed by GenAI models. Moreover, they explore how GenAI can effectively solve optimization problems in energy harvesting wireless networks from numerous perspectives such as channel estimation and relay topology design. Nevertheless, the existing works lack a comprehensive view of prominent applications of DMs in future networks and communications. 

This motivates us to provide a comprehensive review of the applications of DMs to address issues in future networks and communication systems. The contributions of this survey are as follows: 
\begin{itemize}
    \item We first present an extensive tutorial of DMs, including the mathematical backgrounds of the most prominent types of DMs. Subsequently, we demonstrate how DMs are used as optimizers as well as integrated into reinforcement learning and incentive mechanism designs, which have been widely applied
to solve problems in wireless networks
    \item We review and discuss the recent works utilizing DMs in channel modeling and estimation. We also showcase the advantages of DMs in solving channel-related issues. 
    \item We review and discuss innovative frameworks that integrates DMs into signal detection and data reconstruction. 
    \item We review and discuss the latest DM-based approaches for ISAC systems, regarding the most notable problems such as channel and sensing parameter estimation, signal detection and target recognition and interference suppression. 
    \item We review and discuss the uses of DM for improving resource management in edge computing networks and tackling issues, e.g. computation offloading, AIGC service management and designing incentive mechanisms.  
    \item We discuss the enhancements that DMs can provide semantic communication schemes. Subsequently, we explore the recent advancements in leveraging DMs in semantic communication problems such as joint source-channel coding and semantic reconstruction, multi-modal and cross-modal semantic communication, as well as efficient resource-constrained semantic communication. 
    \item We further investigate using DMs for other notable issues including wireless security, spectrum trading, radio map estimation, user association, access control, power control and data collection. 
    \item Finally, we provide insights into the technical limitations of current DM-based approaches and discuss promising research directions related to the integration of DMs into future network and wireless communication systems. 
\end{itemize}

The remainder of the paper is organized as follows. Section \ref{sec:diff_model} introduces the mathematical fundamentals of DMs, as well as their applications in important domains. Section~\ref{sec:DM_for_channel} discusses DM-based frameworks for channel estimation and modeling. Section~\ref{sec:DM_for_signal} investigates using DMs for signal detection and data reconstruction. Section~\ref{sec:DM_for_ISAC} provides insights into applying DMs to solve issues in ISAC systems. Section~\ref{sec:DM_for_EC} provides reviews of integrating DMs into resource management for edge computing networks. Section~\ref{Sec:SemCom} reviews the applications of DMs for semantic communication. Section~\ref{sec:DM_for_other_issues} discusses applications of DMs for other emerging issues. Section~\ref{sec:conclusions} concludes this paper and highlights primary challenges and future research directions. 

\vspace{2em}





\section{Fundamentals of DMs} \label{sec:diff_model}
 By definition, DMs are a family of probabilistic generative models that aim to learn a \textit{diffusion process} for a given dataset, so as to generate new desired samples. In other words, DMs progressively contaminate data with noise, then learn to reverse this process to create new elements. DMs can be either unconditional, often to explore the upper limits of generation capability, or conditional to control the generation results according to our intentions. 
As stated in \cite{vu2024applications}, there are three types of DMs: denoising diffusion probabilistic models (DDPMs) \cite{sohl2015deep, ho2020denoising, nichol2021improved}, score-based generative models (SGMs) \cite{song2019generative, song2020improved} and score stochastic differential equations (SSDEs) \cite{song2020score, song2021maximum}.

In this section, we first provide a brief overview of the DM family. Then, we discuss the mathematical backgrounds of the main types of DMs. Finally, we describe how DMs are integrated into optimizers, reinforcement learning, and incentive mechanisms, which have been widely applied to solve problems in wireless networks.

\subsection{Overview}

Generally, a DM consists of a forward and backward process. In the forward process, a clean sample is sequentially perturbed with noise (it would eventually turn into pure noise in an infinite time-scale). In the backward process, a neural network is trained to gradually remove the added noise distribution in the sample and produce a new clean data distribution.

\subsection{Denoising Diffusion Probabilistic Models (DDPMs)}

A DDPM \cite{sohl2015deep, ho2020denoising, nichol2021improved} utilizes two Markov chains: one corrupts data with noise, the other converts the noise back to meaningful data. The former Markov chain is typically kept simple, with a view to transform any data distribution into a familiar prior distribution. Meanwhile, the latter aims to reverse the former with deep neural networks (DNN)-based transition steps. Generating new data samples is usually conducted by sampling a random vector from the prior distribution and gradually passing it through the reverse chain. 

Formally, the forward Markov process takes a data distribution $x_0 \sim q(x_0)$ and generates a sequence of random variables $x_1, x_2, ..., x_T$ with transition kernel $q(x_t | x_{t - 1})$. Using chain rule and Markov property, we have $        q(x_1, ..., x_T | x_0) = \prod^T_{t = 1} q(x_t | x_{t - 1})$.
DDPMs usually use $q(x_t | x_{t - 1})$ to transform $q(x_0)$ into a tractable prior distribution. One of the most popular choices is to inject Gaussian noise $    q(x_t | x_{t - 1}) = \mathcal{N}(x_t; \sqrt{1 - \beta_t}x_{t - 1}, \beta_t\mathbf{I})$, with $\beta_t \in (0, 1)$ being a predefined hyperparameter. Work in \cite{sohl2015deep} showed that the Gaussian kernel allows marginalization of $q(x_1, ..., x_T | x_0)$, so that the analytical form of the transition kernel can be obtained. Given $\alpha_t \coloneq 1 - \beta_t$ and $\tilde{\alpha_t} \coloneq \prod^t_{s = 1} \alpha_s$, we have $q(x_t | x_0) = \mathcal{N}(x_t; \sqrt{\tilde{\alpha_t}}x_0, (1 - \tilde{\alpha_t})\mathbf{I}).$

With $x_0$ already provided, we can easily generate a sample of $x_t$ by continuously sampling a Gaussian vector $\epsilon \in \mathcal{N}(0, \mathbf{I})$ and applying the  transformation $    x_t = \sqrt{\tilde{\alpha}_t} x_0+ \sqrt{1 - \tilde{\alpha}_t} \epsilon$. As $x_T$ is almost normally distributed when $\tilde{\alpha_T}$ approaches $0$, we get $q(x_T) \coloneq \int q(x_T | x_0) q(x_0) dx_0 \approx \mathcal{N}(x_T; 0, \mathbf{I})$. 

The reverse Markov chain used for reconstruction is typically equipped with a prior distribution $p(x_T) = \mathcal{N}(X_T; 0, \mathbf{I})$ (this distribution is chosen as $q(x_T) \approx \mathcal{N}(X_T; 0, \mathbf{I})$) and a learnable transition kernel $p_\theta(x_{t - 1} | x_t)$, which is expressed by
\begin{equation}
    p_\theta(x_{t - 1}|x_t) = \mathcal{N}(x_{t - 1}; \mu_\theta(x_t, t), \sum_\theta(x_t, t)),
\end{equation}
where $\theta$ denotes the model parameters. The mean $\mu_\theta(x_t, t)$ and variance $\sum_\theta(x_t, t)$ are parameterized by DNNs. With an initial vector $x_T \sim p(x_T)$, this reverse Markov chain allows generation of a data sample $x_0$ by iteratively applying the transitional kernel $x_{t - 1} \sim p_\theta(x_{t - 1} | x_t)$ until $t = 1$. 

\begin{figure}
    \centering
    \includegraphics[width=1.0\linewidth]{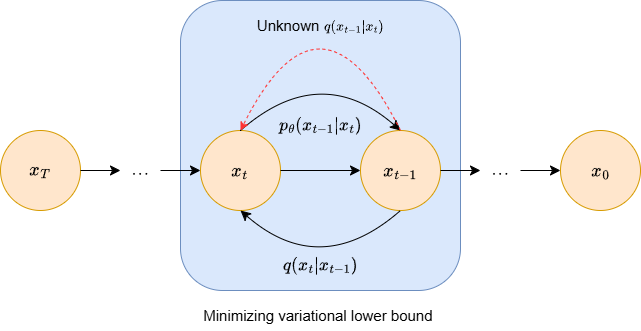}
    \caption{Overview of a typical DDPM. Given the forward Markov chain which adds noise to the original data sample, the aim is to approximate the reverse one with DNNs so that it can remove the noise and produce desired results. This can be done by minimizing the variational lower bound of $x_0$'s log-likelihood.}
    \label{ddpm}
\end{figure}
To train the reverse Markov chain to match the actual time reversal of the forward one, we need to adjust the parameter $\theta$ so that the joint distribution of the reverse Markov chain $p_\theta(x_0, x_1, ..., x_T) \coloneq p(x_T)\prod^T_{t = 1}p_\theta(x_{t - 1} | x_t)$ is a close approximation of that of the forward process $p_\theta(x_0, x_1, ..., x_T) \coloneq p(x_0)\prod^T_{t = 1}p_\theta(x_{t} | x_{t - 1})$. This is achieved by minimizing the Kullback-Leibler (KL) divergence 
\begin{subequations}
    \label{kl-div}
    \begin{align}
       & \text{KL}(q(x_0, x_1, ..., x_T) || p_\theta(x_0, x_1, ..., x_T)) \\
       & = -\mathbb{E}_{q(x_0, x_1, ..., x_T)}[\text{log }p_\theta(x_0, x_1, ..., x_T)] + C \\
       & = \mathbb{E}_{q(x_0, x_1, ..., x_T)}\bigg[
        -\text{log }p(x_T) 
    - \sum^T_{t = 1}\text{log }\frac{p_\theta(x_{t - 1} | x_t)}{q(x_t | x_{t - 1})}
       \bigg] 
      + C \\ 
       & \geq \mathbb{E}[-\text{log }p_\theta(x_0)] + C, \quad \quad \quad \text{(Jensen inequality)}
    \end{align}
\end{subequations}
where $C$ is a constant independent of $\theta$.


DDPMs possess several advantages over contemporary generative models. It ensures much more stable training than GANs \cite{goodfellow2020gan} and is capable of providing high-quality data sampling. However, the slow inference and computational costs of DDPM remains to be addressed. A typical DDPM may need hundreds of denoising steps to achieve decent results, which can be time-consuming and computationally consuming. DDPMs has been applied for network optimization problems \cite{liang2023selection}, resource allocation \cite{wu2024drl} and beamforming \cite{wang2024generative_rl}. 

\subsection{Score-based Generative Models (SGMs)}

SGMs utilize the concept of the score function, defined as the gradient of the log probability density, i.e., $    \nabla_x \log p(x)$ which points in the direction of the steepest ascent of the data density. This property allows SGMs to effectively capture and model complex, high-dimensional distributions without requiring explicit density normalization. A key idea underlying SGMs is to progressively corrupt the data by adding Gaussian noise of increasing intensity, thereby generating a sequence of noisy distributions. Specifically, for each noise level denoted by \(\sigma_t\), a noise-conditional score network (NCSN)  learns to estimate the true score function \(\nabla_x \log p_{\sigma_t}(x)\). This training process is typically accomplished by minimizing the discrepancy between the network's output \(s_\theta(x_t, \sigma_t)\) and the  score of the perturbed data distribution. To achieve this, one commonly employs denoising score matching, with the loss function defined as
\begin{equation}
    \begin{aligned}
    \mathcal{L}(\theta) = \mathbb{E}_{p(x_0)}\,\mathbb{E}_{p(x_t\mid x_0)}\bigg[\lambda(\sigma_t)| s_\theta(x_t, \sigma_t) - \\ 
    \nabla_{x_t} \log p(x_t \mid x_0) |^2\bigg],
    \end{aligned}    
    \label{eq:loss}
\end{equation}
where the Gaussian perturbation kernel is given by $p(x_t \mid x_0) = \mathcal{N}(x_t; x_0, \sigma_t^2 I)$.
In the case of Gaussian noise, the true score function has a closed-form expression:
\begin{equation}
    \nabla_{x_t} \log p(x_t \mid x_0) = -\frac{x_t - x_0}{\sigma_t^2}.
    \label{eq:truescore}
\end{equation}
Substituting Eq.~\eqref{eq:truescore} into Eq. ~\eqref{eq:loss} simplifies the training objective to learning a network that satisfies
\begin{equation}
    \mathcal{L}(\theta) = \mathbb{E}_{p(x_0)}\,\mathbb{E}_{p(x_t\mid x_0)}\left[\lambda(\sigma_t)\left\| s_\theta(x_t, \sigma_t) + \frac{x_t - x_0}{\sigma_t^2} \right\|^2\right],
\end{equation}
where the weighting function \(\lambda(\sigma_t)\) is often chosen as \(\lambda(\sigma_t) = \sigma_t^2\) to balance the contribution of different noise levels. Once the score function is learned, new samples are generated by iteratively refining a noise vector through sampling techniques such as Annealed Langevin Dynamics (ALD). 

SGMs essentially eliminate the need for explicit density normalization, thereby simplifying the training process compared to likelihood-based approaches. Moreover, their flexibility in modeling high-dimensional, complex data and the decoupling of training from the sampling procedure allow for the use of various sampling techniques. However, the iterative sampling process makes SGMs computationally costly and they are not yet fully adapted to discrete data domains such as texts and graphs. SGMs have found numerous applications in channel estimation and modeling, notably in MIMO \cite{arvinte2022mimo, olutayo2023score} and ambient backscatter (AmBC) networks \cite{rezaei2024adversarial}.

\subsection{Score Stochastic Differential Equation (SSDEs)}
DDPMs and SGMs can be scaled to infinite time steps by making the noise perturbation and removal processes solutions to stochastic differential equations (SDEs). This formulation is called \textit{Score SDE} (SSDE) \cite{song2020score}. 
Typically, an SSDE contaminates data with noise through a diffusion process governed by the following stochastic differential equation:
\begin{equation}
    \label{org_sde}
    dx = f(x, t)dt + g(t)d\textbf{w},
\end{equation}
where $f(x, t)$ and $g(x)$ are diffusion and drift functions of the SDE while $\textbf{w}$ is a standard Wiener process. The forward processes in DDPMs and SGMs are both discretizations of this SDE. The work in \cite{song2020improved} shows that the corresponding SDE for DDPMs is $dx = -\frac{1}{2}\beta(t)xdt + \sqrt{\beta(t)}d\textbf{w}$ where $\beta(\frac{t}{T}) = T\beta_t$ as $T \rightarrow \infty$. For SGMs, the SDE is $dx = \sqrt{\frac{d[\sigma(t)^2]}{dt}}d\textbf{w}$, in which $\sigma(\frac{t}{T}) = \sigma_t$ as $T \rightarrow \infty$. 

We use $q_t(x)$ to denote the distribution of $x_t$ in the forward process. The work in \cite{anderson1982reverse} demonstrates that any diffusion process that is expressed in the form of Eq. (\ref{org_sde}) can be reversed by solving the following reverse-time SDE:
\begin{equation}
    \label{rev_time_sde}
    dx = [f(x, t) - g(t)^2\triangledown_x\text{log }q_t(x)]dt + g(t)d\tilde{\mathbf{w}},
\end{equation}
where $\tilde{\mathbf{w}}$ is a standard Wiener process when time flows backward and $dt$ denotes an infinitesimal negative time step. The solutions to the reverse-time SDE are diffusion processes that gradually convert noise to data. Furthermore, there exists an ordinary differential equation (ODE), named \textit{probability flow ODE}, whose trajectories share the same marginals as the reverse-time SDE. This ODE is formulated as:
\begin{equation}
    \label{prob_flow_ode}
    dx = \bigg[f(x, t) - 
    \frac{1}{2}g(t)^2\triangledown_x\text{log }q_t(x) \bigg]dt.
\end{equation}
The probability flow SDE and reverse-time SDE having identical marginals for their trajectories means that we can sample from a data distribution with both of them. Once $\triangledown_x \text{log }q_t(x)$ is known at each time step $t$, we can generate samples by solving the reverse-time SDE (Eq. (\ref{rev_time_sde})) and probability flow ODE (Eq. (\ref{prob_flow_ode})). This can be done with various numerical methods, for example numerical SDE solvers \cite{song2020score, jolicoeurmartineau2021gotta}, annealed Langevin dynamics \cite{song2020improved}, numerical ODE solvers \cite{song2020score, zhang2022fast, song2020denoising_DI} and Markov chain Monte-Carlo \cite{song2020improved}. 

SSDEs are a notable improvement over other types of DMs. They offer a flexible framework that unifies different diffusion processes. Moreover, they can utilize adaptive solvers and faster samplers, thereby ensuring higher sample efficiency. Nevertheless, SSDEs are still quite slow and computationally expensive. There have been only limited applications of SSDEs in wireless communications, such as in channel denoising \cite{mo2025scdm}. 
\begin{table*}[h]
    \centering
    \renewcommand{\arraystretch}{1.5}
    \setlength{\tabcolsep}{8pt}
    \begin{tabular}{l p{4.5cm} p{4.5cm} p{4.5cm}}
        \toprule
        \textbf{Model Type} & \textbf{Nature} & \textbf{Advantages} & \textbf{Drawbacks} \\
        \hline 
        \midrule
        \multirow{3}{*}{\textbf{DDPM}} 
        & Discrete-time Markov chain that progressively adds and removes noise. 
        & - High-quality image generation. \newline
          - Strong likelihood-based training. \newline
          - Theoretical foundation in variational inference. 
        & - Slow sampling due to long Markov chain steps. \newline
          - Requirement of many denoising steps for good results. \\
        \midrule
        \multirow{3}{*}{\textbf{SGM}} 
        & Use of the function (gradient of log probability) of noisy data to guide generation.
        & - Flexible noise scheduling. \newline
          - More efficient sampling than DDPM. \newline
          - Score-matching based training. \newline 
        & - High computational costs. \newline
          - Requirement of careful noise conditioning. \\
        \midrule
        \multirow{3}{*}{\textbf{SSDE}} 
        & Continuous-time formulation using SDEs to model noise evolution. 
        & - Unification of DDPM and SGM. \newline
          - Providing interpretation through stochastic processes. \newline
          - Enabling adaptive sampling strategies.
        & - Requirement of solving complex SDEs. \newline
          - Higher computational cost for large-scale models. \\
        \bottomrule
    \end{tabular}
    \caption{Comparison of different types of DMs}
    \label{tab:diffusion_models}
\end{table*}

\subsection{Diffusions Models As An Optimizer Solution}

Network optimization is of the most challenging tasks in wireless communications systems, and DMs (and generative models as a whole) have demonstrated their promising advantages over discriminative models in this task \cite{liang2024diffusion}. 

We formulate a typical network optimization problem as  $\text{min}_{{y} \in \mathcal{Y}} f({x}, {y}), x \in \mathcal{X}$ where ${x}$ and ${y}$ represent input and output parameters, respectively, $f \rightarrow \mathbb{R}$ is the objective function, and $\mathcal{X}$ and $\mathcal{Y}$ are the domain and feasible region, respectively. Discrete-time DDPM has been proposed to learn the distribution of high-quality solutions $p({y}_0 | {x})$, with ${y}_0$ representing the original data \cite{liang2024diffusion}. As the noise perturbation process does not depend on $x$, we simplify this term to $p(y_0)$. The dataset $\mathcal{D}$ consists of $n$ data pairs, each in the form of $({x}, y)$. For the noise perturbation process, the original data is gradually corrupted by Gaussian noise with mean and variance controlled by noise factor $(\alpha_t)^T_{t = 1}$. With an uncorrupted training sample $y_0 \sim p(y_0 | x)$, the noisy samples $y_1, y_2, ..., y_T$ is obtained through the forward Markov process
\begin{equation}
    \label{forward}
    p(y_t | y_{t - 1}) = \mathcal{N}(y_t; \sqrt{\alpha_t} y_{t - 1}, 
    (1 - \alpha_t)\mathbf{I}), \forall t \in 1, .., T,
\end{equation}
where $T$ represents the number of diffusion steps and $I$ is the identity matrix with the same dimension as $y_0$. 

To conveniently acquire corrupted data $y_t$ from any time step, we leverage the recursive feature of Eq. (\ref{forward}) using the formula $ p(y_t | y_0) = \mathcal{N}(y_t; \sqrt{\overline{\alpha}_t} y_{t - 1}, (1 - \overline{\alpha}_t)\mathbf{I})$ where $t \sim \mathcal{U}({1, ..., T})$ and $\overline{\alpha}_t$ is the product of $a_i$ ($i$ ranges from $1$ and $t$). Specifically, noise $\epsilon_t$ is sampled from a standard normal distribution $\mathcal{N}(0, \mathbf{I})$ for any sample $(\mathbf{x}, \mathbf{y})$ in $D$. A sample $\mathbf{y}_t$ is randomly chosen ($t$ is sampled from $U({1, ..., T})$) and injected with noise as $\mathbf{y}_t = \sqrt{\overline{\alpha}_t}\mathbf{y}_0 + \sqrt{1 - \overline{\alpha}_t} \epsilon_t$. The denoising process principally reverses the corrupted sequence by trying to estimate the true posterior distribution with a Gaussian process parameterized with $\theta$ as $    p_\theta(\mathbf{y}_{t - 1} | \mathbf{y}_t) = \mathcal{N}(\mathbf{y}_{t - 1}; \mu_\theta(\mathbf{y}_t, \mathbf{x}, t), \sum_\theta(\mathbf{y}_t, \mathbf{x}, t))$.

To handle conditional guidance from $\mathbf{x}$ during the denoising process, classifier-free guidance \cite{ho2022classifier} can be used, which jointly trains conditional and unconditional models. Specifically, a hyperparameter $p_{\rm{uncond}}$ is introduced into the training phase to constrain the extent to which the model is trained unconditionally. The resulting prediction noise would be $\overline{\epsilon}_\theta(\mathbf{y}_t, \mathbf{x}, t) = (1 + \omega)\epsilon_\theta(\mathbf{y}_t, \mathbf{x}, t) - \omega\epsilon_\theta(\mathbf{y}_t, t)$,
where $\omega$ dictates the intensity of conditional guidance. The reconstruction function, based on $\overline{\epsilon}_\theta$, is given by $\mathbf{y}_{t - 1} = \frac{1}{\sqrt{\alpha_t}}(\mathbf{y}_t - \frac{1 - \alpha_t}{\sqrt{1 - \overline{\alpha}_t}}\overline{\epsilon}_\theta) + 
    \frac{1 - \overline{\alpha}_{t - 1}}{1 - \overline{\alpha}_t}\epsilon$.


\subsection{DMs for Reinforcement Learning Algorithms}
\label{sec:DM_for_DRL}


DMs have emerged as a powerful tool in reinforcement learning (RL) due to their ability to model complex, multimodal distributions and iteratively refine outputs through stochastic denoising processes. 

\begin{figure}
    \centering
    \includegraphics[width=0.8\linewidth]{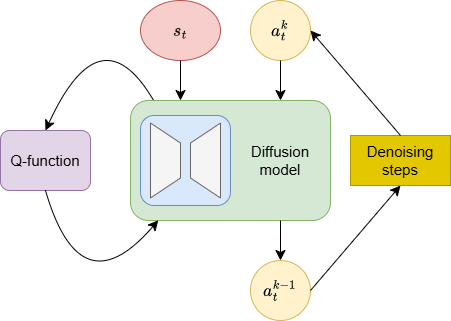}
    \caption{DM as the policy itself. The sampling target is the action given the observed state, usually guided by the Q-function.}
    \label{dm_policy}
\end{figure}

Typically, a DM can act as the policy \cite{estruch2023idql} (see Fig (\ref{dm_policy})). Additionally, it can also serve as a data synthesizer or planner \cite{janner2022planning}, where the sampling target is the trajectories \cite{ajay2022diffuser} or parts of them. In the data synthesizer case, both real and synthetic data are used for downstream policy optimization. The RL objective is typically to maximize the cumulative reward: $    J(\pi) = \mathbb{E}_{\tau \sim \pi}\left[\sum_{t=0}^{T} \gamma^t r(s_t, a_t)\right],$
where $\pi$ denotes the policy, $\gamma \in [0,1)$ is the discount factor, $r(s_t, a_t)$ is the reward at time $t$, and $\tau$ is a trajectory $(s_0, a_0, s_1, a_1, \dots)$. DMs generate candidate actions by iteratively refining noisy samples. The reverse diffusion process in the context of action generation can be modeled as: $    p_\theta(a_{t-1} \mid a_t, s_t) = \mathcal{N}\left(a_{t-1}; \mu_\theta(a_t, s_t), \Sigma_\theta(a_t, s_t)\right)$, 
where $\mu_\theta(a_t, s_t)$ and $\Sigma_\theta(a_t, s_t)$ are learnable functions guiding the denoising process based on $s_t$ and the noisy $a_t$.

To integrate the RL objective directly, a loss function is defined to promote the generation of actions with high expected returns: $\mathcal{L}_{RL}(\theta) = \mathbb{E}_{s \sim \mathcal{S},\, \epsilon \sim \mathcal{N}(0,I)}\left[- Q\Big(s, \mathcal{D}_\theta(\epsilon, s)\Big)\right]$
where $\mathcal{D}_\theta(\epsilon, s)$ denotes the reverse diffusion process that maps noise $\epsilon$ (conditioned on state $s$) to an action, and $Q(s,a)$ estimates the expected return of taking action $a$ in state $s$. This loss encourages the DM to generate actions that lead to higher rewards. In addition, the standard DM loss is used to train the denoising network as $\mathcal{L}_{DM}(\theta) = \mathbb{E}_{a \sim \mathcal{A},\, \epsilon \sim \mathcal{N}(0,I)}\left[\|\epsilon - \epsilon_\theta(a, s)\|^2\right],$
where $\epsilon_\theta(a, s)$ is a neural network that predicts the noise added to action $a$ given state $s$. Minimizing this loss ensures effective denoising, contributing to robust exploration and exploitation.

For considerable advantages, DMs have been used to enhance RL algorithms in various wireless communication domains, including resource allocation \cite{wu2024drl, darabi2024diffusion}, UAV-assisted IoT system optimization \cite{liang2024uav} and UAV communication optimization \cite{zhang2024multi}.  

\subsection{DMs for Incentive Mechanisms}

DMs have become the prominent choice for incentive mechanism design, for example Stackelberg game, auction theory and contract theory thanks to their ability to generate entirely unseen data samples across various domains.  
\subsubsection{Game Theory}
DMs are increasingly deployed to solve complex game-theoretic equilibria. Unlike conventional solvers limited to low-dimensional strategies, diffusion-based approaches parameterize players' action spaces as learnable noise distributions. For instance, SGMs can be leveraged to approximate Nash equilibria \cite{babichenko2018incentive} in imperfect-information games through iterative strategy refinement: $    \mathbf{a}_t^{i-1} = \mathbf{a}_t^i - \eta \nabla_{\mathbf{a}_t^i} \mathcal{L}_{\text{NE}}(\mathbf{a}_t^i, \mathbf{s}_t)$,
where $\mathbf{a}_t^i$ represents player actions at denoising step $i$, and $\mathcal{L}_{\text{NE}}$ encourages equilibrium convergence given game states $\mathbf{s}_t$.

\subsubsection{Contract Theory}

DMs can be used in contract-based frameworks to encourage information sharing \cite{du2023ai}
among users. The contract is first randomly generated as Gaussian noise and goes through multi-step denoising, whose final output is a nearly-optimal contract design that maximizes predefined optimization objectives. Specifically, the contract design policy is represented as a reverse process of a DM as $    \pi_\theta(\mathbf{c}|\mathbf{e}) = p_0(\mathbf{c}^{0:N}|\mathbf{e}) = 
\mathcal{N}(\mathbf{c}^N; 0, \mathbf{I}) \prod^N_{i = 1}p_\theta(\mathbf{c}^{i - 1}|\mathbf{c}^{i}, \mathbf{e})$,
where $\mathbf{c}$ is the contract to be designed given the environment and $\mathbf{e}$ are the environment states.

Similarly, DDPM are utilized to generate contracts as a means to mitigate the adverse selection problems when IoT devices share sensing data for digital twin (DT) construction~\cite{wen2025sustainable}. This framework can also be employed for incentivizing edge AIGC service providers in 6G IoT networks \cite{wen2024diffusion}. Other notable examples can be found in vehicular network design \cite{zhong2024generative} and AIGC service resource allocation \cite{ye2024optimizing}.

\subsubsection{Auction Theory}
Auction design benefits from DMs synthesizing revenue-optimal mechanisms without closed-form solutions. For example, DMs are used for designing combinatorial auctions \cite{zhang2023optimal} where the model progressively denoises reserve prices and allocation rules to maximize expected revenue as $\max_{\boldsymbol{\theta}} \mathbb{E}_{\mathbf{b} \sim p_{\text{bid}}}[R(\mathcal{D}_{\boldsymbol{\theta}}(\mathbf{b}))]$ and 
    $\mathcal{D}_{\boldsymbol{\theta}} = \text{ReverseDiffusion}(\boldsymbol{\epsilon}, \mathbf{b})$,
with $\boldsymbol{\epsilon}$ as initial noise and $\mathbf{b}$ as bid distributions. Additional work, such as \cite{li2020incentive} and \cite{fang2023multiunit}, further explores incentive-compatible diffusion auctions in various auction settings.

\subsection{Summary}
Table.~\ref{tab:diffusion_models} summarizes each type of DM's strengths and weaknesses. Specifically, there are three main types of DMs, namely DDPMs, SGMs and SSDEs, each with their own unique attributes. DDPMs offer stable training and high-quality outputs but suffer from slow inference due to many denoising steps. Given these advantages, DDPMs are used to solve a wide range of problems, including semantic communication, channel modeling and data generation for ISAC systems. SGMs improve efficiency in modeling high-dimensional continuous data by leveraging score matching and do not require explicit likelihoods. Thus, they are leveraged for tasks in which explicit density modeling is difficult, such as massive MIMO signal detection and channel estimation. Finally, SSDEs provide a unified version of SGMs and DDPMs with even greater generation capability, but come with a higher computational cost due to complex numerical solvers. As a result, SSDEs are utilized for semantic communications and tasks with extremely low-SNR conditions.

\section{DMs for Channel Modeling and Channel Estimation}
\label{sec:DM_for_channel}
To achieve high data rate in next-generation wireless networks, accurate Channel State Information (CSI) plays a pivotal role in enabling tasks such as coherent data recovery, adaptive beamforming, and precise localization in next-generation wireless systems \cite{dovelos2021channel}. Obtaining accurate CSI is challenging due to dynamic, complex wireless channels (fading, shadowing, interference) and the high dimensionality/computational overhead in wideband/MIMO systems, hindering real-time estimation.

Understanding and mitigating channel effects is paramount, requiring channel modeling and estimation. Modeling creates a mathematical/statistical representation of signal transformation, capturing phenomena like path loss, fading, and noise \cite{jiang2020channel}. Channel estimation uses the channel model to infer real-time channel parameters (CSI) from received signals.
Traditional channel estimation methods, such as MMSE and Maximum-Likelihood estimators, have long been the mainstays of wireless communications. These methods often rely on simplified channel models and require precise knowledge of noise statistics, which are often unavailable in practice. Furthermore, their performance degrades significantly in non-stationary and low signal-to-noise ratio (SNR) environments \cite{liu2014channel}. In recent years, machine learning (ML) has emerged as a promising alternative. DL-based approaches, like ChanEstNet \cite{liao2019chanestnet}, can learn complex channel patterns directly from data, eliminating the need for explicit channel models.

DMs offer a compelling alternative for channel modeling and estimation, potentially overcoming the limitations of traditional and DL-based methods. DMs excel at capturing complex data distributions by gradually denoising a random noise signal. This inherent capability to model intricate statistical relationships makes them well-suited for characterizing the stochastic nature of wireless channels. Specifically, the inherent characteristics of DMs make them highly advantageous for wireless communication systems \cite{yang2023diffusion, du2024enhancing}. DMs provide \textit{robustness to wireless noise}, being specifically designed to operate in noisy environments and thus resilient to prevalent interference. DMs offer \textit{flexibility in modeling complex distributions}, allowing them to learn non-Gaussian and non-stationary channel behaviors that traditional models often simplify. Their \textit{data generation capabilities} enable the creation of realistic channel samples, which can significantly enhance the robustness and generalization of channel estimation algorithms through data augmentation. The denoising process within DMs also acts as an \textit{implicit regularizer}, effectively preventing overfitting and improving the generalization performance of channel estimation. Lastly, \textit{sampling flexibility} from DMs allows the number of samples to be adjusted based on available computational resources and desired accuracy.

\subsection{Channel Modeling}

Recent advancements in wireless communications have seen a surge in the applications of DMs for various channel modeling tasks \cite{xu2025fully,lee2024generating,sengupta2023generative}. Particularly, \cite{xu2025fully} employs conditional DMs to generate typical channel representations for efficient modulation and coding scheme (MCS) selection in dynamic environments. A two-tiered hierarchical reinforcement learning (HRL) method is proposed for co-frequency MIMO-OFDM transmission, optimizing precoding matrices and rank indicators. This HRL approach, which uses a transformer encoder for subcarrier correlation, allows the higher level to select the MIMO spatial layer and the lower level to determine precoding for coordinated base stations. A conditional diffusion model is then utilized to generate a representative over all time-varying subframe channels. Specifically, a forward process progressively introduces random Gaussian noise into the time-varying channel data via a Markov chain, while a reverse diffusion process is performed to generate a channel representative of the original data distribution.
Simulations show their feedback-free approach from two base stations improves UE throughput by 13\% over single-base station closed-loop spatial multiplexing and achieves 94\% performance of a heuristic code-book iteration method, presenting a promising diffusion design, though real-world data is needed.

With respect to the specific application of MIMO channel generation, \cite{lee2024generating} utilizes conditional denoising diffusion implicit models (cDDIM) to generate synthetic channels from positional data. In particular, the score function, designed with Langevin dynamics \cite{song2019generative}, generates desired samples. It is learned from data via denoising score matching, which does not require the underlying channel distribution \cite{vincent2011connection}. This method is to create rescaled, perturbed channel versions (with Gaussian noise) for which the conditional score is easily computed analytically.
This technique generates highly accurate synthetic channel data by conditioning the model on user locations. This effectively expands dataset, addressing the limitations of sparse measurements.
The method's effectiveness is proven by its superior performance over noise addition and GANs in training channel compression and beam alignment, especially with limited data. 
It accurately determines the dominant beam index 40\% of the time, and is within one index 67\% of the time.
Although progressive distillation and consistency models are investigated, the perfect results are not achieved like standard DMs. This difference poses a significant challenge for real-time applications for further investigation. 
Using different approaches, \cite{sengupta2023generative} uses latent diffusion to learn channel distributions from sparse MIMO data, demonstrating DM versatility and bypassing the need for explicit domain expertise.
DM iteratively refines Gaussian noise to accurately represent channel characteristics, overcoming GAN limitations like training instability and mode collapse to generate diverse, high-fidelity channel samples mirroring the true distribution.
This model demonstrates adaptability by learning from a large urban macro-cellular environment and accurately modeling a distinct urban micro-cellular setting using only 5-10\% of its data, proving its robustness to real-world variations.
\begin{figure}[t] 
	\centering 
	\includegraphics[width =1\linewidth]{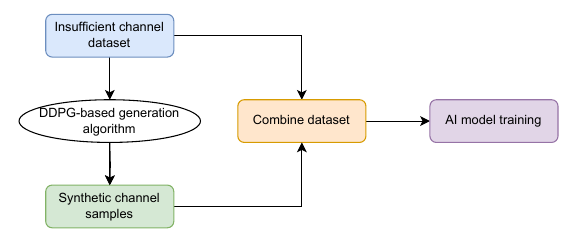} 
	\caption{DDPM-based data augmentation process flow \cite{xu2023denoising}. DDPM's strong generative capability allows us to learn channel distribution and create synthetic channel data. Combining this with the real dataset effectively closes the performance gap.} 
	\label{fig:DDPM_DA} 
\end{figure}

DMs are able to generate synthetic samples, and thus they can used to combat data scarcity issues of the channel modeling \cite{zhang2024denoising,xu2023denoising}. 
To combat data scarcity issues, \cite{zhang2024denoising} use a DDPM-based digital twin (DT) to generate synthetic channel data that reflects real-world distributions (measured by Kullback-Leibler divergence), enabling effective DL control with limited samples and dynamic conditions. 
This involves data creation, diffusion, and denoising, utilizing a U-Net architecture with time embeddings and attention for better feature handling.
The DT-based channel generator assisted framework is constructed to improve the performance of data-driven DL algorithms for sensing channel estimation and target detection. The provided results demonstrate that the DDPM based DT effectively optimizes the performance of a DnCNN for channel estimation and enhances target detection accuracy within an SNR range of -15dB to 10dB. To validate the scheme's effectiveness across diverse impacts and applications, the scalability of the current system must be considered.
\cite{xu2023denoising} leverages DDPM for data augmentation to combat limited training data, generating synthetic channel samples as depicted in Fig. \ref{fig:DDPM_DA}.
The CSI feedback task is used as a test case to measure the effectiveness.  Simulations reveal a substantial performance boost in terms of NMSE with small training datasets started from 500 samples, which gradually decreased as the amount of training data increased over 3500 samples. 
Further testing on various channel types is needed to validate the proposed scheme's robustness and adaptability.

\begin{figure}[t] 
	\centering 
	\includegraphics[width =1\linewidth]{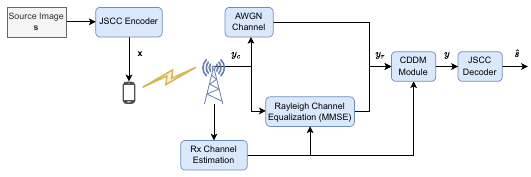} 
	\caption{Architectural overview of the joint CDDM-JSCC system \cite{wu2023cddm}. CDDM after equalization predicts and eliminates channel noise to enhance performance. Its forward diffusion process, based on the received signal's conditional distribution (Rayleigh fading/AWGN), facilitates a unique training algorithm and a sampling method for noise elimination. CDDM module is then integrated into a JSCC-based semantic communication system for wireless image transmission, with its output fed to the JSCC decoder for image recovery.} 
	\label{fig:CDDM_JSCC} 
\end{figure}

To mitigate channel noise in wireless communications, \cite{wu2023cddm} proposes Channel Denoising DMs (CDDM). Designed for Rayleigh and AWGN channels, CDDM aims to recover the original source signal by learning its distribution using MMSE equalizer and normalization-reshape techniques. Thus, CDDM is trained by optimizing a closed-form variational bound on the negative log-likelihood, derived using the Rao-Blackwellized method.
CDDM is trained with a noise schedule designed for wireless channels, allowing it to effectively eliminate channel noise through a specific sampling algorithm in the reverse process. This makes CDDM suitable for semantic communication systems employing joint source-channel coding (JSCC). The architectural configuration of the CDDM and JSCC integrated system is illustrated in Fig. \ref{fig:CDDM_JSCC}.
The experiments demonstrate that the combined CDDM and JSCC system gains 1.06 dB in PSNR at 20 dB SNR over Rayleigh fading compared to JSCC alone. 
However, the current approach is limited to the tested noise models, and its effectiveness with other noise types remains uncertain.
For low-density parity-check (LDPC) decoding, \cite{guan2024joint} proposes a novel RNN-DDPM technique that jointly enhances channel equalization and decoding. This method enables joint denoising by exchanging LDPC parity information between the modules, effectively combining the benefits of parity checks and the DM's Markov process.
The proposed LDPC decoding method achieves a 0.2dB to 0.5dB improvement in BER compared to standard normalized min-sum LDPC decoding in both AWGN and Rayleigh channels. 
Further research should focus on integrating channel estimation into the LDPC decoding method and optimizing the full interaction between the receiver's core modules.

Beyond conventional channel samples, DMs apply in a variety of contexts, including space-air-ground integrated networks and radio map construction \cite{zhang2024generative,wang2024radiodiff}. For space-air-ground integrated networks (SAGIN), \cite{zhang2024generative} addresses concerns surrounding generative AI's application in the SAGIN environments by presenting a comprehensive review and case study. Their work demonstrates how generative AI models can be integrated into SAGIN, leveraging their ability to generate data and enhance decision-making. This confirms the potential of generative AI within this network architecture.
In \cite{wang2024radiodiff}, RadioDiff, a Diffusion Model (DM) for accurate path loss estimation, is introduced. It uses decoupled diffusion and Adaptive Fourier Transforms (AFT) to capture environmental dynamics, representing static and dynamic features separately and employing AFT for sensitivity to rapid changes. 
RadioDiff surpasses existing methods in accuracy, structural similarity, and peak signal-to-noise ratio. However, reducing inference time and exploring advanced DM techniques are vital for generating detailed environmental features from limited data.

\subsection{Channel Estimation}

A range of research efforts focuses on developing DMs for channel estimation suitable for diverse applications in mobile systems \cite{bhattacharya2025successive,xu2024generative,zeng2024dmce,ma2024diffusion}.
For instance, \cite{bhattacharya2025successive} introduces a new algorithm that combines successive interference cancellation (SIC) with DMs to improve the channel estimation and data detection. This method iteratively refines the estimated channel and source data, using a channel gain-based SIC decoding order to determine which channel portions to estimate. The key idea of this approach is to utilize DMs to estimate the structure of wireless channels by focusing on individual submatrices. By processing submatrices, this replicates massive MIMO channel conditions, and the DMs achieve optimal results when the channel matrices are nearly full rank.
The findings show a clear performance advantage for the proposed algorithm, with superior NMSE and SER compared to baseline methods, observed in both low-rank and full-rank channels with $10$, $40$, and $70$ symbols. In \cite{ma2024diffusion}, the authors introduce a novel approach to channel estimation by leveraging DM-based posterior sampling. DMs are utilized to learn the score functions of the posterior channel data distribution, where posterior sampling within the reverse processes of DDPM and Denoising Diffusion Implicit Model (DDIM) is employed to reconstruct the true channel response. 
The performance of the proposed DDPM- and DDIM-based solutions is validated by comparing them against the Linear Minimum Mean Squared Error estimator and a Score Matching and Langevin Dynamics-based channel estimators, using Clustered Delay Line-C channel data. The findings demonstrate the superior performance of the proposed DM-based estimator, where DDIM-based solution exhibits improved performance while significantly reducing computational complexity compared to the DDPM-based approach. 
While the current work focuses on a simple network, accurately assessing real-world systems requires considering systems with more users to capture complex impact factors and diverse patterns.

In the same context, \cite{xu2024generative} employs conditional DMs for wireless data generation and DM-powered DRL for communication management, focusing on DM-driven channel generation to enhance MIMO channel estimation, prediction, and feedback.
Leveraging precise DeepMIMO data, this research shows that conditional DMs can generate realistic channel representations and handle unexpected variations in complex wireless environments.
To address unreliable channels and task demands, DM-based communication management strategies are emerging. Further work is needed to optimize, including accelerating sampling, integrating with mobile edge computing, and exploring model-driven learning interactions, for full DM-driven communication in mobile networks.

For semantic communication, \cite{zeng2024dmce} proposes a multi-user system that fuses traffic scene images by transmitting extracted semantic features over a MIMO channel, then reconstructed at the receiver, and enhanced by a DM-based channel enhancer over challenging environments.
By utilizing semantic vectors from compressed images integrated channel equalization, this approach gains multi-user image decoding accuracy. DMCE at the receiver refines CSI for effective equalization and reduced interference, leading to more accurate signal recovery and improved semantic image reconstruction.
This approach significantly improves object segmentation in low SNR conditions. Validated on a traffic scenario dataset \cite{ha2017mfnet}, it achieves over 25\% mIoU compared to benchmarks. Incorporating multi-modal information could further enhance performance.

SGMs are utilized in various works to enhance MIMO channel estimation and data detection \cite{zilberstein2024joint,arvinte2022mimo}. Specifically, \cite{zilberstein2024joint} proposes a DM algorithm with SGMs for joint massive MIMO detection and channel estimation, integrating discrete symbol and learned channel priors.
This method addresses blind inverse problems by using a diffusion process to approximate the joint posterior distribution of symbols and channels, enabling maximum a posteriori estimation through sampling.
While numerical tests show a significant performance advantage, surpassing established baselines, i.e., LASSO \cite{venugopal2017channel}, fsAD \cite{bhaskar2013atomic}, L-MMSE \cite{nayebi2017semi}, L-DAMP \cite{metzler2017learned}, by orders of magnitude at SNRs above 15 dB, this analysis is limited to uplink scenarios and does not address downlink operation.
Unlike the above study, \cite{arvinte2022mimo} introduces a novel unsupervised, probabilistic method for downlink MIMO channel estimation that minimizes pilot symbol requirements using a score-based approach for posterior sampling.
This work presents a precise mathematical formula for robust SISO channel estimation via posterior sampling. Experiments show it achieves accurate MIMO channel estimation (up to 64x256) with just 25\% pilot density. It also surpasses compressed sensing in scalability, offering reduced computations and lower latency for larger systems.
However, research is needed on adapting pre-trained models to new environments with minimal training data, which would expand their applications.

Leveraging DDPMs, \cite{zhou2024generative} proposes a MIMO channel estimation technique that uses denoising diffusion generative models and learned channel priors for accurate recovery via posterior inference, even with low-resolution analog-to-digital quantization.
To boost wireless transmission reliability, DMs are trained using an unsupervised Stein's unbiased risk estimator, enabling effective learning from noisy data without requiring impractical ground truth channel information.
The proposed estimator offers high-fidelity channel reconstruction for real-time, scalable ultra-massive MIMO systems, reducing latency by 10x and pilot overhead by 50\% compared to baselines such as EM-GM-AMP \cite{vila2013expectation}, BLMMSE \cite{li2017channel}, VAE \cite{fesl2024channel}.
Using a channel simulator for data generation, while done, fails to capture real-world channel complexities.
Using similar approaches, \cite{fu4973643conditional} introduces a conditional denoising diffusion-based channel estimation (CDDCE) method for MIMO-OFDM systems, designed to handle fast time-varying channels. By employing an inverse fast Fourier transform to extract initial channel features, followed by iterative refinement using a trained U-Net, CDDCE effectively estimates channels under complex fast-fading conditions. 
This scheme demonstrably enhances channel estimation and signal detection, achieving a strong performance-complexity trade-off. CDDCE scheme surpasses traditional and advanced DL methods in NMSE and BER across a wide SNR range (5-30 dB), showcasing its ability to learn channel statistics. However, the additional work should explore techniques like knowledge distillation to reduce computational overhead and address scenarios with limited training samples.

Focusing on the lightweight model to reduce complexity, the authors in \cite{fesl2024diffusion} introduce a new MIMO channel estimator utilizing DMs. 
Initially, DMs generate latent variables by progressively adding Gaussian noise in a forward process. The reverse process, also a Markov chain, reconstructs the data by reversing these noisy steps. Since the reverse transitions are difficult to calculate directly, they are approximated by using neural network. This network is trained using variational inference, leveraging the known distributions from the forward process. A key contribution is a streamlined CNN designed for efficient inference, minimizing complexity and memory overhead while maintaining strong performance.
This work proves that the DM-based channel estimator asymptotically converges to the minimum MSE optimal conditional mean estimator. Furthermore, the DM exhibits a substantial performance improvement of up to 5 dB SNR over estimators, i.e., GMM \cite{koller2022asymptotically}, GMM Kron \cite{fesl2022channel}, score-based model \cite{arvinte2022mimo}, demonstrating robustness to SNR mismatch. 
The current model is tailored for a specific channel condition and system parameter, thus its viability in channels with high variability remains to be validated through further study.

\begin{figure}[t] 
	\centering 
	\includegraphics[width =1\linewidth]{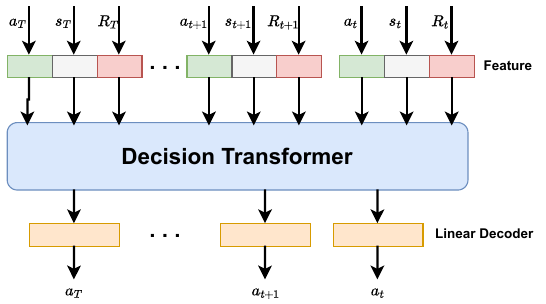} 
	\caption{Architectural schematic of the Decision Transformer model \cite{zhang2025decision}.} 
	\label{fig:DecisionTransformer} 
\end{figure}

To address the challenge of limited network coverage, reconfigurable intelligent surfaces (RIS) or intelligent reflecting surfaces (IRS) are widely employed \cite{huang2022reconfigurable,Qingqing2024}. These surfaces, equipped with advanced hardware containing multiple reflecting elements, reflect base station signals to users in obstructed areas \cite{swindlehurst2022channel}. Numerous studies explore channel estimation in these RIS-assisted communication systems \cite{zhang2025decision,zhang2024decision,tong2024diffusion1}. In particular, \cite{zhang2025decision} introduces a Diffusion-Enhanced Decision Transformer (DEDT) to improve beamforming in RIS-assisted communication. This framework combines a DM for better CSI acquisition with a decision transformer for adaptive beamforming, leading to improved accuracy and environmental adaptability. The visual explanation of the Decision Transformer's design is shown in Fig. \ref{fig:DecisionTransformer}.
To address the difficulty of accurately generating CSI, DM using partial information is proposed. By conditioning DM on a subset of channel data, the model learns the spatial relationship between partial and complete CSI. This enables the generation of accurate, real-time CSI. While the reverse denoising phase provides probability distributions, calculating them is complex due to the need for joint probabilities across all steps. Similar to \cite{fesl2024diffusion}, a neural network is used to approximate these distributions. In particular, the framework of decision transformer  is developed to enable adaptive decision-making based on historical data, limiting retraining requirements.
The simulation studies reveal a 7.5\% performance gain for the proposed method, which employs a Transformer-based model in DEDT, compared to RL algorithms. 
In pursuit of the same goal, \cite{zhang2024decision} presents the Diffusion-Decision Transformer (D2T) to improve beamforming in IRS-assisted MISO systems. D2T utilizes DM for channel estimation, offering reduced complexity and more accurate real-time channel recovery than traditional methods, thereby improving efficiency, especially in diverse conditions.
A two-stage approach is used to ensure reliable decision models across channels: offline Decision Transformer (DT) pre-training on diverse data, followed by online few-shot fine-tuning on new channels for adaptation without full retraining.
The pre-trained D2T method demonstrates excellent zero-shot performance in unseen channel conditions. With minimal fine-tuning data, it outperforms traditional RL methods by 6\% and converges 3x faster. Its performance is close to the ideal limits of the DT method with perfect CSI. Diverse user signals could negatively impact performance due to the system's lack of multi-user consideration in both of the above works.
To mitigate phase noise, \cite{tong2024diffusion1} proposes a DM-based RIS channel estimation technique to enhance robustness against received noise and RIS phase noise. Their method involves a forward phase corrupting the true channel with Gaussian noise and a reverse phase iteratively reconstructing it using a likelihood-trained U-Net that integrates gradient descent of the RIS phase.
Simulations demonstrate significant improvements, with NMSE reduced by over $3.2$ dB and a $3.74$ dB gain over phase noise-ignorant approaches. However, the current evaluation is limited to a single user, requiring further investigation with a broader range of users. 

Tackling the specific characteristics in RIS-aided massive MIMO systems, the authors in \cite{liu2022channel} address cascaded channel estimation by employing an unsupervised DM. 
This DM operates in the spectral domain of the degradation matrix, learning the channel structure without labeled data and with a loss function independent of the degradation model. During sampling, the received signal is incorporated as a condition, steering the iterative channel recovery.
Simulations demonstrate their scheme outperforms Bayesian methods, achieving 2-3 dB lower NMSE with less pilot overhead.
However, the study is limited to uplink scenarios, requiring downlink analysis for full validation.

\subsection{Lessons Learned}
DMs are rapidly emerging as a powerful toolset for both channel estimation and modeling in wireless communications. By leveraging conditional denoising diffusion, several studies can generate realistic channel representations tailored to specific system settings, such as UE positions or channel properties, enabling efficient MCS selection in dynamic environments. In scenarios with limited data, DDPMs facilitate channel data augmentation, particularly valuable for MIMO systems \cite{zhou2024generative}. Furthermore, DMs address channel noise reduction, as demonstrated by CDDM, enhancing signal integrity \cite{wu2023cddm}. For channel estimation, SGMs are utilized to improve detection and leverage posterior sampling. The applications of DMs extend to MIMO-based channels, space-air-ground integrated networks, and radio map applications, tackling data scarcity and noise reduction.

A significant limitation across many studies utilizing DMs for channel estimation and modeling lies in their reliance on simulated data, which inherently fails to capture the intricate and often unpredictable dynamics of real-world wireless environments.
Consequently, the performance gains demonstrated in simulation may not translate directly to practical scenarios. Furthermore, the prevalent use of simplified network settings in these studies overlooks the complexities of real-world systems, such as varying user densities, interference patterns, and environmental factors, all of which can significantly impact the efficacy of diffusion-based approaches.

\section{DMs for Signal Detection and Data Reconstruction}
\label{sec:DM_for_signal}
Signal processing and data analysis continuously seek better methods for extracting information and restoring data, especially when dealing with complex noise and limited datasets. Two central tasks are signal detection and data reconstruction. Signal detection focuses on identifying specific signals or discrete information within noisy data, like determining transmitted symbols in a communication system ~\cite{liu2021deeptransfer}. In contrast, data reconstruction aims to restore high-quality data from degraded, incomplete, or compressed versions across various formats like audio and images ~\cite{adam2020sampling}. Essentially, detection is about making decisions, while reconstruction is about generating a high-fidelity representation of the data. Traditional methods, such as algebraic least squares (ALS) and minimum mean square error (MMSE), are often limited by complex noise, data scarcity, and the demand for high-fidelity results~\cite{gao2015lowcomplexity, cogna2017semisupervised}. This can lead to distortions or instability during training. DMs offer a powerful alternative. They work by first systematically adding noise to data and then training a network to reverse this process, iteratively denoising the data to reconstruct it. This noise-to-signal approach provides robust data recovery, performs exceptionally well in low SNR conditions, and generates high-quality samples, effectively overcoming many limitations of older techniques.

\subsection{DMs for Signal Detection}
Recent advancements in DMs introduce innovative generative frameworks for radio signal detection, especially in complex settings like massive MIMO. Their unique design methodology -- a forward noising process followed by an iterative denoising reversal -- progressively refines signals from noise. This offers robust recovery, enhanced performance in challenging conditions, and high-fidelity estimates, addressing shortcomings of prior methods. The work in~\cite{he2024massive} introduce approximate diffusion detection (ADD) to improve signal detection at low SNRs by performing iterative reverse process to stochastically refine the noisy received signal towards valid symbols. Based on simulation results, ADD significantly outperforms traditional baselines such as ALS and MMSE detection in terms of BER, particularly at low SNRs. In spite of demonstrating faster convergence compared to maximum-likelihood detection, it incurs higher latency due to iterative sampling, despite its design aiming to balance performance and complexity.
Building upon a regular framework, the work in~\cite{wang2025erasing} aims to surpass traditional ML-based estimation with a denoising DM based on stochastic differential equations (SDEs). This innovative approach mathematically links SNR to the diffusion timestep, thus allowing the benefit of SNR adaptability without extensive retraining and effective Gaussian noise elimination. Evaluations on BPSK/QAM show lower symbol error rates than classical ML-based methods (reliant on known noise models) and other neural network approaches while maintaining $\mathcal O(n^2)$ complexity. Further work might explore scalability or more diverse noise.

\begin{figure}[t] 
	\centering 
	\includegraphics[width =1\linewidth]{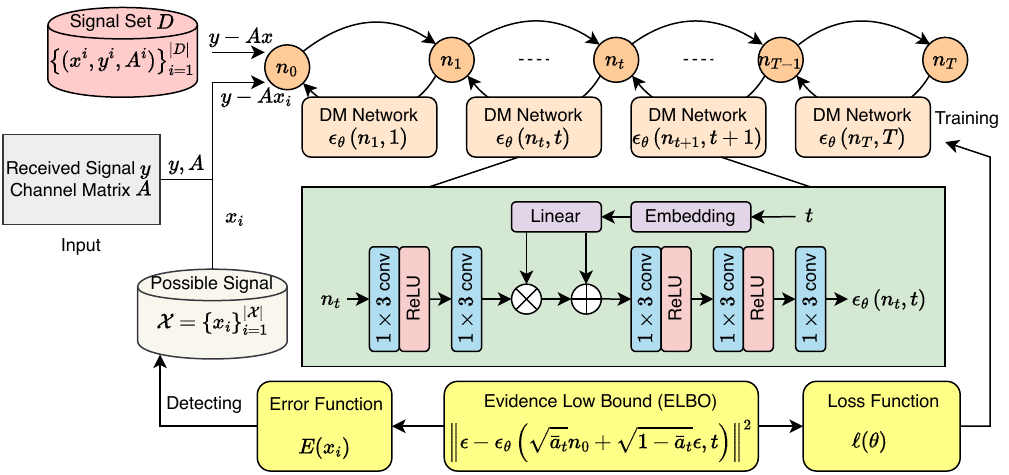} 
    \caption{The MLEDD framework~\cite{zhao2024signal} first trains a diffusion model to learn the noise distribution, and then detects the transmitted signal by using the trained model to find the candidate signal that minimizes a calculated error value.}
	\label{fig_MLEDD} 
\end{figure}

Further extending the application of DMs to scenarios with unknown interferences, the work in~\cite{olutayo2023score} proposes a score-based generative model for MIMO detection. The primary motivation is to eliminate the need for prior knowledge of noise statistics besides overcoming architectural constraints faced by previous neural network-based approaches. By leveraging SDEs to map an unknown noise distribution in the received signal into a known Gaussian latent space, the score-matching approach in this work offers greater flexibility in neural network architecture design. This strategy bypasses invertibility constraints of methods like maximum normalizing flow estimate (MANFE) and generalized approximate message passing (GAMP), accordingly achieving near-optimal ML detection under arbitrary noise conditions. Simulation results demonstrate the superiority of the proposed model over both GAMP and MANFE across a variety of noise types, including Gaussian, colored Gaussian, Nakagami-m, and impulsive noise. While the method benefits from reduced parameter complexity, it suffers a higher runtime due to ordinary differential equations computations.
For near-field communication with unknown noise, the authors in~\cite{zhao2024signal} introduce the maximum-likelihood estimation diffusion detector (MLEDD) with the training and detection processes illustrated in Fig.~\ref{fig_MLEDD}. MLEDD's core technical innovation is its ability to learn complex noise distributions directly from data during training. This diffusion-based design offers the remarkable benefit of surpassing traditional methods, especially when noise distributions are intractable or unknown, thus leading to superior performance in such conditions. Extensive evaluation results reveal MLEDD's superior BER performance compared to several baselines—including matched filters, regular detection networks, and MANFE across various SNR regimes. While these results demonstrate MLEDD's potential in effectively handling complex noise, its computational cost should be optimized for real-time near-field applications.

\subsection{DMs for Data Reconstruction}
Leveraging their inherent ability to learn complex data distributions and reverse degradation processes, DMs adeptly reconstruct high-fidelity data from various corrupted, incomplete, or compressed inputs across different modalities, consequently proving highly effective for demanding signal and image restoration and generation tasks. The following subsections detail the applications of DMs for reconstructing radio symbols and signals, speech and audio, and images.

\subsubsection{Radio Symbol/Signal Reconstruction}

\begin{figure}[!t] 
	\centering 
	\includegraphics[width =1\linewidth]{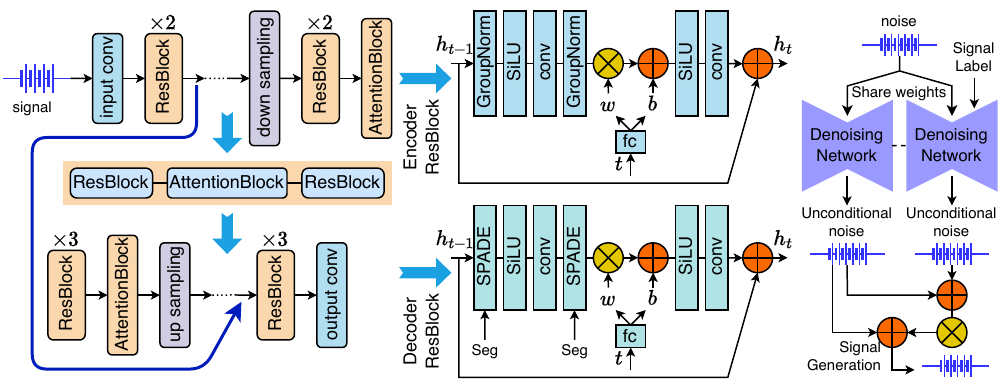} 
    \caption{The SEI-DM architecture~\cite{zha2023sei} consists of an encoder-decoder and a diffusion model. It generates RF signals using classifier-free guidance, which merges conditional and unconditional noise predictions during the denoising-based sampling process.}
	\label{fig_sei_dm} 
\end{figure}
DMs are emerging as transformative tools in radio signal processing, offering novel methods to boost communication reliability and efficiency. Their core design—a forward noising stage followed by a reverse denoising or generation stage—uniquely equips them to learn intricate data distributions and produce high-fidelity signals, particularly in adverse environments.
To enhance symbol transmission reliability, the work in~\cite{letafati2023probabilistic} proposes optimizing symbol generation using DDPMs. Uniquely, their DDPM-based approach mimics receiver reconstruction to adaptively shape symbols by SNR. The key benefit of this diffusion methodology is its inherent denoise-and-generate capability, leading to a closer alignment between transmitted and received symbols. Compared to the methods directly optimizing constellation distributions, this approach increase a threefold mutual information, but suffering high complexity for real-time implementation.

To address limited training data for automatic modulation recognition (AMR), the authors in~\cite{chen2024data} develop a conditional DM for dataset enhancement. The technical approach involves using the conditional DM to iteratively denoise random noise into realistic modulated signals, effectively generating high-confidence synthetic samples for training dataset enhancement. The inherent benefit of diffusion is its strong generative capability, thus producing diverse and high-quality signals that capture the nuances of different modulations.
For efficient, goal-oriented communication, the work in~\cite{wijesinghe2024diff} introduces Diff-GO, a diffusion approach with low-dimensional noise space mapping and local generative feedback to enhance message recovery and prioritize relevant data. Diff-GO outperforms traditional semantic communication frameworks in bandwidth management and semantic integrity preservation, but being limited by high complexity of the feedback mechanism in dynamic scenarios.

Motivated by challenges in RF signal generation for specific emitter identification (SEI) in low-resource settings, the authors in~\cite{zha2023sei} develop SEI-DM, illustrated in Fig.~\ref{fig_sei_dm}, to iteratively denoise and refine emitted signal characteristics. 
SEI-DM's key innovations are a one-dimensional U-Net for capturing temporal dependencies in RF signals and classifier-free guidance for conditional signal generation, which strengthens feature separability.
Extensive evaluations report a significant accuracy improvement, compared with GANs, at low SNR regimes, consequently confirming SEI-DM's ability to mitigate data scarcity, suppress overfitting, and enhance recognition under strong noise.
To improve data reconstruction in cell-free massive MIMO downlink systems suffering from hardware imperfections and interference, the work in~\cite{letafati2024diffusion} deploys denoising diffusion implicit models (DDIMs). This approach formulates data transmission as forward diffusion, then uses reverse diffusion to remove impairments, accordingly conducting effective operation with incomplete CSI. This work yields a flexible trade-off between computational complexity and reconstruction quality through adjustable hyperparameters.

\begin{figure}[t] 
	\centering 
	\includegraphics[width =1\linewidth]{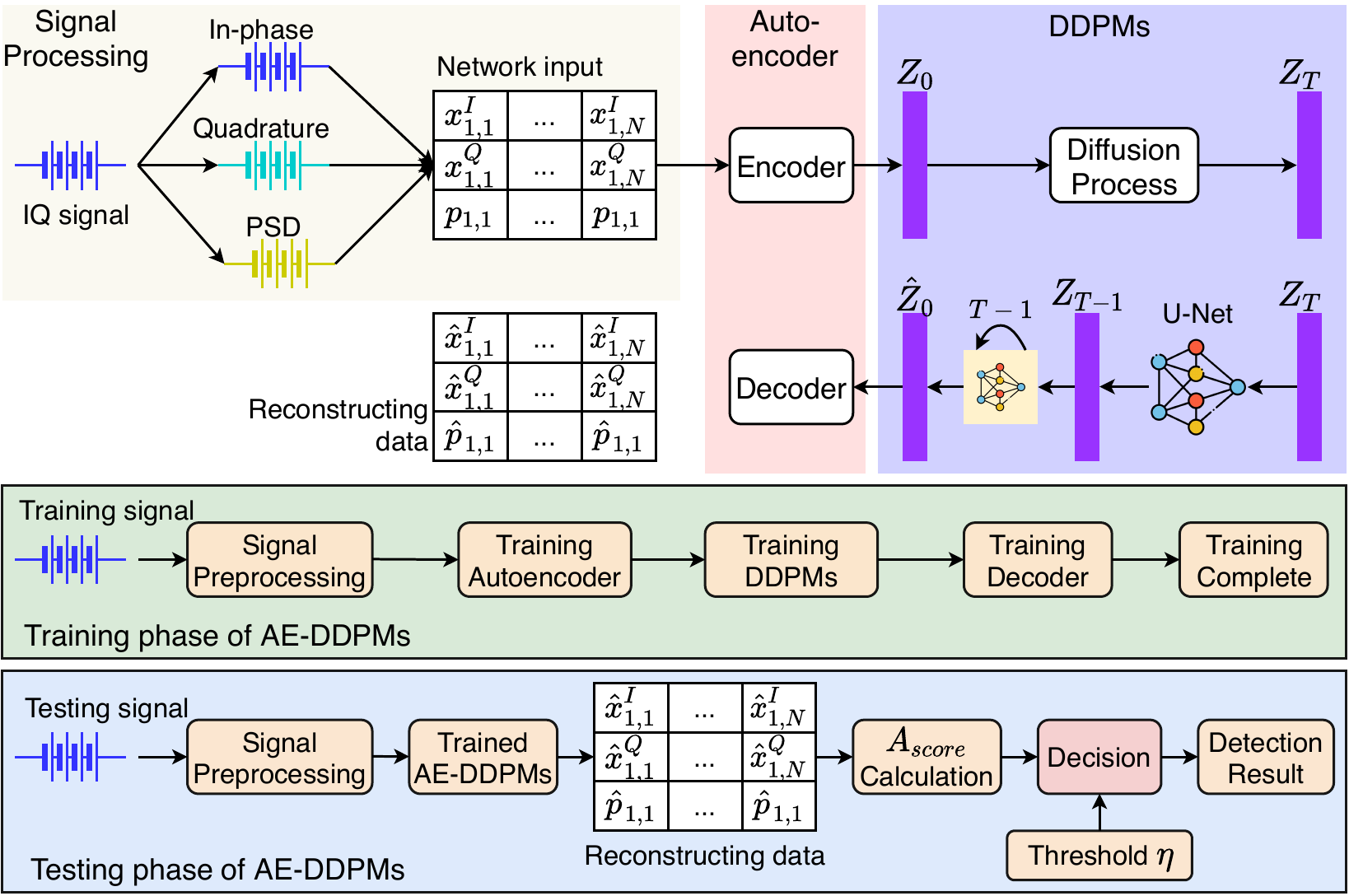} 
    \caption{The AE-DDPMs framework~\cite{zeng2023radio} integrates an autoencoder for feature compression with a DDPM for feature denoising. The model learns normal signal characteristics during training to perform signal reconstruction and classification.}
	\label{fig_ae_ddpm} 
\end{figure}


For radio signal anomaly detection, the work in~\cite{zeng2023radio} develop autoencoder-DDPMs (AE-DDPMs), an advanced diffusion-oriented autoencoder framework as illustrated in Fig. \ref{fig_ae_ddpm}. In a lower-dimensional latent space, AE-DDPMs are beneficial in both complexity and accuracy by accurately identifying anomalies via reconstruction errors from the learned normal signal distribution. Experiments show AE-DDPMs outperform SOTA anomaly detection methods, such as matched-filter detection and encoder-GAN, by $12$ dB at low SNRs with fewer diffusion steps.
In summary, DMs leverage their distinct generative and refinement capabilities to advance wireless communications in symbol optimization, data augmentation, efficient transmission, RF signal generation, and complex MIMO reconstruction, tackling data scarcity and interference for more robust, intelligent radio systems.

\subsubsection{Speech Reconstruction}
In the last decades, DMs are significantly advancing speech enhancement and audio processing by leveraging their unique generative capabilities to address complex reconstructive tasks.
Motivated by the need to overcome distortions from conventional DNNs in 3D speech enhancement, the authors in~\cite{chen2024two} introduce a two-stage system, in which DMs are deployed as backend refiners after neural beamforming, accordingly the DM’s generative nature, trained on clean speech, distinctly enhances naturalness and accuracy in restoring distorted speech components. Experimental results demonstrate consistent improvements in word error rate and quality metrics across various datasets and beamformer architectures compared to baseline beamformer outputs or traditional post-filters, thus highlighting the DM's strong generalization.
To ensure semantic coherence in audio communication despite severe degradation, the work in~\cite{grassucci2024diffusion} proposes a novel framework with the capability of uniquely formulating audio restoration as an inverse problem, in which a conditional latent diffusion model~\cite{rombach2022high} is aptly upgraded for simultaneous denoising and inpainting from lower-dimensional representations. As a remarkable achievement, this diffusion approach offers robust generation of semantically coherent audio even with substantial data loss or noise. Compared to unspecified existing audio restoration methods, this approach obtains superior Frechet Audio Distance and SNR, accordingly a new standard for resilient audio communication.
To overcome the limitations of single-microphone DMs and independent channel processing in multi-microphone MIMO speech enhancement, the authors in~\cite{kimura2024diffusion} develop a multistream score-based generative method. This approach uniquely performs joint analysis of multi-microphone signals using a DM conditioned on these signals, a key benefit being efficient and robust modeling that preserves crucial spatial cues (e.g., inter-microphone differences) against varying array geometries. Integrating techniques like weighted prediction error dereverberation further boosts its effectiveness, thus considerably outperforming single-microphone DMs or MIMO systems that process channels independently, especially in noisy, reverberant conditions.
In summary, these studies highlight the transformative role of DMs in speech reconstruction. By harnessing their generative capabilities for spatial filtering, semantic audio communication, and multi-microphone enhancement, DMs are driving the development of more robust and high-quality speech technologies.

\subsubsection{Image Reconstruction}
Beyond audio, DMs offer innovative solutions for image reconstruction and synthesis, enhancing quality and restoring degraded visuals. To address reliable image reconstruction in extreme conditions, Cdiff~\cite{letafati2024conditional} is a conditional diffusion model designed for robustly recovering severely degraded images. By directly integrating degraded received images into the denoising framework, Cdiff leverages a key advantage of DMs for robust image recovery and reconstruction. Cdiff achieves over $10$ dB improvement compared to traditional DNN-based receivers (which often lack such generative refinement) on the NVIDIA Sionna platform, accordingly showing promise for diverse applications.
To address the perception-distortion trade-off in image transmission, the work in~\cite{yilmaz2024high} integrates a DDPM into a deep joint source-channel coding (DeepJSCC) framework, which enables lower-resolution image transmission over wireless channels by effectively leveraging the DM’s generative power for high-perceptual quality refinement at the receiver. This DM-driven framework significantly outperforms standard DeepJSCC (without DMs) and GAN-based JSCC methods across various SNRs and bandwidths, consequently enabling high-quality recovery under challenging conditions.

In hyperspectral imaging, the authors in~\cite{liu2023diverse} present a DM-based approach for synthesizing diverse remote sensing images, accordingly addressing the limited spectral diversity issue. As a key contribution, this approach employs a latent-space diffusion process, where a conditional vector quantized GAN first compresses data; the DM then efficiently operates in this latent space, thus enhancing spectral diversity and spatial consistency from color inputs. Relying on benchmark results on the IEEE grss\_dfc\_2018 dataset~\cite{xu2019advanced}, this approach, benefiting from the DM's ability to handle high-dimensional noise and enhance output diversity, outperforms non-DM deep models that incorporate advanced techniques such as response-function-based guidance and hybrid attention mechanisms.
For spectral super-resolution, the authors in~\cite{chen2024spectral} propose a spectral-cascaded DM. Unlike previous single-shot techniques, this model is notably featured by a coarse-to-fine diffusing flow to effectively model and learn complex spectral relationships, further enhanced by image condition mixture guidance. 
The iterative refinement learning strategy enhances the model’s performance over conventional CNN- and GAN-based approaches, thus yielding improvements across metrics such as root mean square error (RMSE) and peak signal-to-noise ratio (PSNR).
These studies highlight DMs' transformative potential in image processing by tackling challenges like transmission noise, perception-distortion trade-offs, and spectral data complexities.

\subsection{Lessons Learned}
In summary, DMs advance signal detection and data reconstruction with superior performance over traditional methods, especially in low-SNR and complex noise environments. These models are proficient in data recovery, perceptual quality, and generative tasks like data augmentation and refinement. However, they suffer from high computational costs and latency from iterative sampling along with complex training and design~\cite{olutayo2023score}.
DMs are particularly recommended for tackling challenges where traditional methods often fall short. They are highly effective for signal detection in low SNRs and complex/unknown noise environments~\cite{zhao2024signal}, high-fidelity data reconstruction from corrupted or incomplete data (e.g., radio, speech, and images)~\cite{he2024massive}, and generative tasks such as data augmentation, high-quality content creation, and generating novel samples~\cite{zha2023sei}.
To effectively exploit DMs, their core properties like learned generative priors and stochasticity should be leveraged in model design for enhanced accuracy and practicality. Notably, conditional modeling improves DM performance by allowing customization to specific inputs or attributes. Additionally, hybrid architectures (e.g., DMs with autoencoders or as refiners) offer complementary gains, accordingly boosting performance for high-dimensional or multi-stage processing~\cite{zeng2023radio}. Furthermore, shaping DM architectures and training strategies to specific domains and data characteristics is essential, involving selecting appropriate network backbones and preserving critical data features. Finally, balancing performance with practical constraints (such as computational demands and inference latency) is essential, thereby requiring strategies to optimize sampling or operate in learned latent spaces for real-time or resource-limited applications.


\section{DMs for Integrated Sensing and Communication}
\label{sec:DM_for_ISAC}

Integrated Sensing and Communication (ISAC) is a cutting-edge technology that merges wireless communications (e.g., data packet transmission) and environmental sensing (e.g., extracting target information using radar-like waveforms) functions within a shared spectrum and hardware platform, enabling simultaneous data transmission and perception  \cite{lu2024integrated}. This integration enhances spectrum efficiency, reduces hardware costs, and supports various applications, including autonomous vehicles, smart cities, and 6G networks \cite{Nuria2024}.  However, amidst noise and interference, realizing ISAC networks encounter significant challenges in terms of complex signal processing and resource management, with key issues: communication channel and sensing parameter estimation, signal detection and target recognition, data generation and reconstruction, and interference suppression. Traditionally, such problems can be addressed using the following common methods, but they have some limitations. First, Kalman filtering and Least Squares (LS) estimations struggle with non-linear channels, joint signals, and dynamic environments, leading to inaccurate parameter estimations. Second, matched-filtering and clustering algorithms are limited by noise, clutter, and low SNR factors, thereby decreasing detection accuracy. Third, compressive sensing and error correction codes render high computational complexity and are quite sensitive to signal interference, hindering reliable reconstruction. Lastly, beamforming and successive interference cancellation become ineffective in dense, dynamic scenarios with overlapping signals, causing performance degradation.   

Recently, DMs (DMs) have been known as a new alternative deployed at the receiver to process incoming signals or in base-band processing units to model complex data distributions and adapt to dynamic conditions, offering superior performance. In what follows, we will review how DMs address the aforementioned issues through their advanced technical capabilities.

\subsection{DMs for Communication Channel and Sensing Parameter Estimation}
Channel and Sensing Parameter Estimation is the foremost task in ISAC for determining channel characteristics (like fading and interference) using various techniques (e.g., pilot, blind/semi-blind, LS and MMSE), as well as sensing parameters (such as range and velocity) using Kalman filters. This ensures accurate signal interpretation for real-time applications. However, the variability in downlink and uplink transmissions in ISAC networks complicates estimating channel attributes and signal directions departing from and arriving at the target, despite advancements in learning frameworks that utilize data-driven methods \cite{respati2024survey} to train ISAC transceivers in offline and exploiting them for online prediction. These learning-based models depend heavily on the quality and quantity of data collection, making them less suitable for dynamic ISAC applications.
%

\begin{figure}[!t]
    \includegraphics[width=1.023\linewidth]{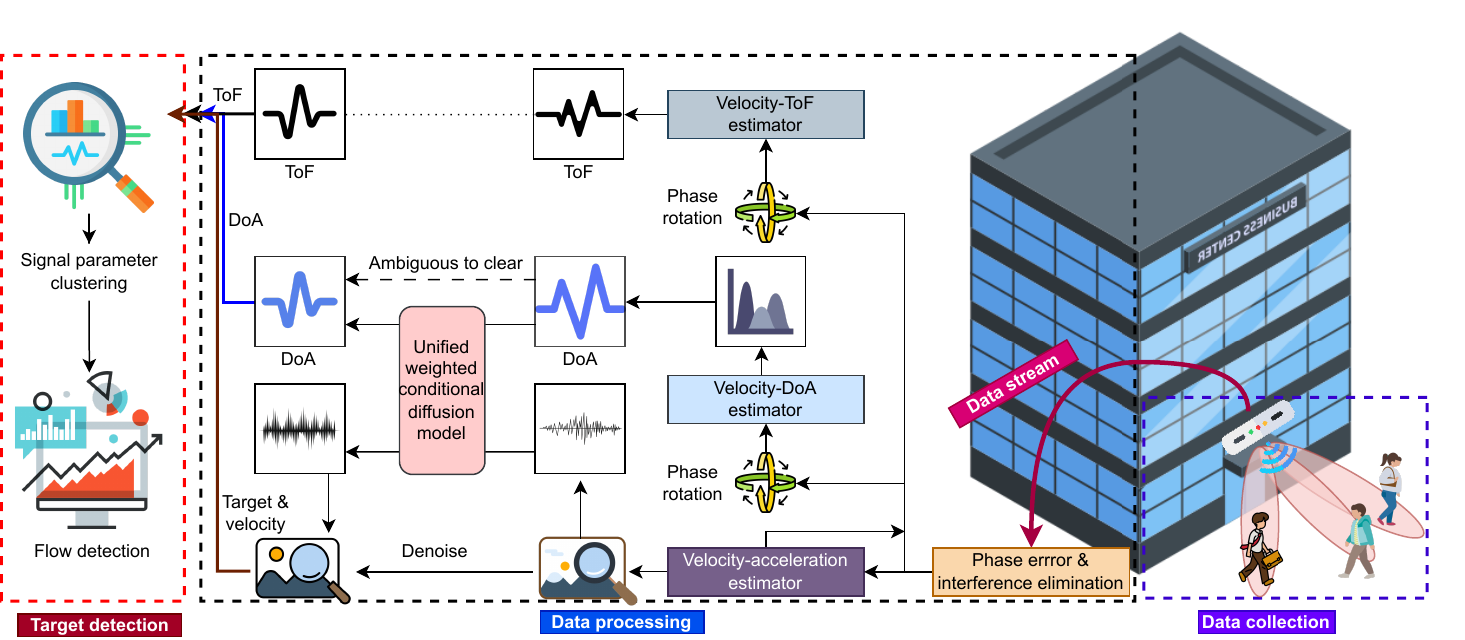}
    \caption{Illustration of DMs for human detection in near-field scenarios \cite{wang2024generative_ai}.}
    \label{fig:wang2024generative}
\end{figure}

In this context, the DM proposes a solution by treating the forward process as a Markov chain and adding diffusion noise to transform ground truth inputs into a standard Gaussian distribution \cite{wang2024generative_sensing}. The reverse process, trained via variational inference with parameterized Gaussian transitions, maps observed channels or angles to time steps within the Markov chain, refining estimates and enhancing denoising. However, practical communication situations impact specific DM refinement. 
%
%
For instance, in near-field scenarios, \cite{wang2024generative_ai} dealt with challenges in the relationship between signal spectrum and antenna spacing adjustments by introducing a weighted conditional DM that generates direction of arrival spectra and enhances CSI measurement denoising. The core idea of this approach (see Fig.~\ref{fig:wang2024generative}) is to constrain the DM by a condition and compel it to generate a desirable spectrum with reduced noise, thereby supporting human follow detection.   
%
%
In massive MIMO, \cite{xu2025brownian} incorporates a Brownian bridge process in the DM, mapping LS estimation to the ground-truth channel and addressing controllability issues in the diffusion process. With proper SNR fine-tuning from \(-10\) dB to \(10\) dB, this approach significantly reduces the number of required diffusion steps and outperforms LS and linear MMSE techniques ($>10$ times) and DL-based methods like DnCNN, U-Net, and cGAN.
Also, DMs can be extended to weather radar forecasting \cite{he2024diffsr}, where a two-stage diffusion method generates detailed radar composite reflectivity by transforming satellite data into image-level data, followed by a conditional DM that denoises Gaussian noise using radar sensing data in batches.

\subsection{DMs for Signal Detection and Target Recognition}
Signal Detection and Target Recognition are defined as the processes of identifying, classifying, and distinguishing communication and sensing signals or environmental targets (e.g., objects, activities, or vital signs) from overlapping, noisy, or cluttered signals. Overlapping ISAC signals make it difficult to distinguish between them, which often causes inefficient resource utilization. Besides, limited IoT network resources also result in sparse and insufficient ISAC signal samples, which undermines the accuracy and reliability of detection and recognition procedures.

As wireless data demand grows for IoT tasks, traditional approaches like matched-filtering and clustering face limitations. Advanced methods, including deep learning classifiers (e.g., CNNs), sparse signal recovery, and time-frequency analysis (e.g., short-time Fourier transform), offer improvements but struggle with computational complexity and dynamic environments. Recently, some research efforts are investigating DM in ways of generative AI for wireless data augmentation to enhance quality and availability to better meet IoT needs \cite{wen2024generative}. This approach shows promise for boosting the performance and robustness of ISAC systems in complex and resource-constrained environments.

For instance, a novel DM-based ISAC scheme introduced in \cite{jiang2024electromagnetic} utilizes sensing channels from ISAC signal reflections to create detailed target point clouds, enabling accurate electromagnetic property sensing. In \cite{xu2024conditional}, a DM-based data augmentation method addresses limited WiFi data by generating additional samples conditioned on activity classes, improving WiFi sensing performance in various scenarios. Focusing on the outdoor environment problem (e.g., detecting terrain, buildings, and infrastructure), \cite{wang2024generative_rf} leverages ISAC signals transformed into image sensors, with a CNN-based Segmentation DMs generating refined barrier maps from sparse power samples, achieving better accuracy and perceptual quality. 

The authors in \cite{sun2024sar} design an SAR recognition method that enhances performance under few-shot conditions by generating diverse features from limited samples using DDPM and similarity-based scattering calculations. In \cite{wu2024robust}, a fully differentiable radar-camera framework is proposed to tackle radar sensing sparsity, enhancing 3D object detection by aligning radar point clouds with images and employing DDPM in a Rao-Blackwellized manner. To mitigate privacy risks from model memorization, \cite{wangprivacy} presents a hybrid training method for DMs that treats WiFi data defensively against membership inference attacks. This approach reduces the attack success rate from $97\%$ to $72\%$ through pre-training and selective application of SSDEs for noise reduction. 

\subsection{DMs for Data Generation and Reconstruction}

After detecting and classifying ISAC signals, the next step in improving the ISAC system is to develop advanced learning algorithms to optimize these signals \cite{respati2024survey}. Data Generation and Reconstruction are crucial for creating dual-purpose signals that convey communication data and facilitate sensing, while also recovering data from noisy or sparse samples to ensure integrity for both functions. Yet, technical challenges persist in extracting key features from data signals across time, radio frequency, and spatial domains. Wireless sensing data collection is time-consuming and costly, and the data varies due to randomness and factors such as transceiver locations, environmental conditions, device types, waveforms, frequency ranges, and protocols. Consequently, a dataset collected at a specific time or for a particular application may not generalize to other times or broader settings, limiting its utility.

One alternative to this dilemma is to use DM to enhance data training, with DDPM serving as an optimal strategy for generating new data. For instance, a two-step approach proposed in \cite{chi2024rf} utilizes a novel time-frequency diffusion theory to leverage RF signal characteristics and employs a hierarchical diffusion transformer for practical implementation. In \cite{qosja2024sar}, DDPM is applied in synthetic aperture radar (SAR) imaging, focusing on specific networks and parameters for both conditional and unconditional scenarios. Another example is the use of conditional DDPM in \cite{bian2024wi} to convert CSI data into multi-channel images, addressing class imbalance and achieving over 94\% classification accuracy. Moreover, the authors in \cite{wang2024aigc_2,wang2024aigc,wang2025aigc} introduce an activity class conditional DM to generate synthetic RF data across various technologies while exploiting conditional DMs to enhance SAR imaging input signals. On addressing network topology visualization challenges, \cite{wang2024empowering} presents a GAI-based framework that uses conditional DMs to convert noise into graph structures, optimizing routing and resource allocation via a reward function.

On the other hand, SGM offers a strategic solution for high-quality data reconstruction. In \cite{wen2024radio}, a two-stage SGM synthesizes high-quality segmentation by encoding foreground patterns across multiple views before processing them in the diffusion pipeline. In \cite{li2024diffgait}, the DiffGait model augments gait data for mmWave recognition using a combination of an autoencoder and adversarial discriminator. Lastly, the method discussed in \cite{lan2024constrained} showcases DM's effectiveness in generating high-resolution images from low-frequency inputs, significantly improving resolution and clutter removal in deep ground penetrating radar images.

\subsection{DMs for Interference Suppression}
As mentioned in the prior section, exploiting ISAC systems can benefit ubiquitous communication and high-precision sensing but it raises serious concerns about how to vacate interference from ISAC signals (i.e., self-interference, mutual interference, clutter, and crosstalk) \cite{xu2024interference}, even when they can be well classified. Besides, the mutual dependence of waveform design, transceiver configuration, and signal processing algorithms highly compromises the function of ISAC in terms of throughput and sum rate while reducing the radar's adaptive detection range. This adversely affects positioning accuracy and diminishes coverage capabilities. Traditionally, such a problem can be resolved using the interference suppression, avoidance, and exploitation methods \cite{niu2024interference}, but encountering model availability and dimensionality challenges. Likewise, exploiting data-driven approaches also questions the deep coupling of various resources, such as costly feature collection, limited bandwidth, time, power, computing capability, etc. Consequently, there is a strong need for advanced interference management solutions to adapt to the rapidly evolving ISAC systems. 

\begin{figure}[!t]
    \centering
    \includegraphics[width=\linewidth]{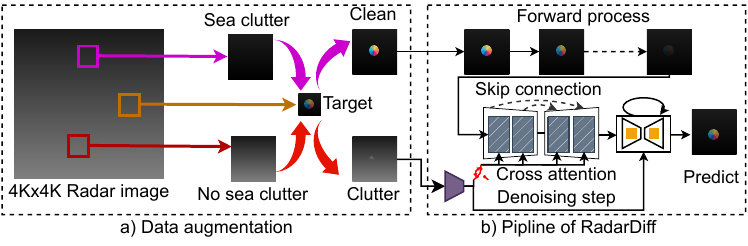}
    \caption{RadarDiff methods in \cite{si2024radardiff} for sea cluttering suppression: a) the suppression of sea clutter is regarded as a task in image-to-image translation through data augmentation, and b) Radar images with sea cluttering are used as the input, U-Net-based diffusion model predicts the combination of sea cluttering and adding noise.}
    \label{fig-si2024radardiff}
\end{figure}

Several DNNs focusing on signal priors through Bayesian algorithms have emerged, such as radar interference mitigation using deep unfolding \cite{overdevest2023signal}. These methods rely on accurate signal priors, a strength of expressive generative models like DMs. For instance, RadarDiff \cite{si2024radardiff} employs DMs to enhance radar target detection by treating sea clutter suppression as an image translation task (see Fig.~\ref{fig-si2024radardiff}), significantly outperforming traditional clutter-reduction and GAN-based methods. Qualitative results show improvements in learned perceptual image patch similarity (LPIPS of $0.7569$), clutter suppression ratio (CSR of $-0.9562$), and mean average precision (mAP of $0.979$), while reducing the Frechet inception distance (FID) to $9.32$. Additionally, \cite{overdevest2024model} proposes joint-conditional probabilistic DMs for automotive radar interference removal, and \cite{lin2024diffusion} introduces an unsupervised anti-jamming method that requires extensive training iterations. Notably, \cite{wu2024diffradar} presents a method for improving low-angle resolution and multipath interference through a contour encoder and semantic decoder for mmWave sensing, enhancing radar representations for accurate scene reconstruction.

On another front, the authors in \cite{luan2024diffusion} propose using the SSDE method to handle LiDAR ghost points and enhance the sparse mmWave radar point clouds. In the meantime, the authors in \cite{zhang2024towards} consider combining the strength of DMs and cross-modal learning approach to predict LiDAR-like point clouds from paired raw radar data. For more general cases, the authors in \cite{huang2024v2x} develop a multi-modal denoising diffusion module for a cooperative LiDAR-4D radar fusion pipeline to robust weather-3D detection. This module treats weather-robust 4D radar features as a condition for DMs. Through testing with the V2X-R dataset, the proposed LiDAR-4D radar fusion pipeline can efficiently suppress noisy LiDARs. 


\section{DMs for Resource Management in Edge Computing Networks}
\label{sec:DM_for_EC}

Edge computing (EC) has emerged as a promising paradigm ~\cite{xu2021edge} which enables the end users to offload their computation tasks to the nearby edge servers (ESs) via wireless channels~\cite{lin2019computation}. 
Due to the time-varying wireless channels and ECs' dynamic resource availability, 
it is challenging to efficiently manage the EC resources to achieve the best task execution performance.
The advanced learning algorithms like deep reinforcement learning (DRL) have been widely adopted to address these challenges 
because of their ability in capturing the complex state space representations. 
However, the DRL algorithms often require extensive training with a lot of online interactions with the environment, 
especially when the system is complex with high-dimensional state and action spaces. 
Meanwhile, the offload DRL training approaches may lead to the poor-performance policies 
due to the function approximation errors on out-of-distribution actions~\cite{wang2022diffusion}. 
As discussed earlier, the remarkable ability of the DMs in simulating the plausible environment dynamics and synthesizing the high-reward trajectories has inspired their use to enhance the sample efficiency and exploration of the DRL algorithms. Thus, many existing studies have proposed various approaches to integrate the DMs into the DRL-based resource management frameworks for the EC systems. In this section, we will review these existing works 
which can be divided into the following three categories depending on their objective: computation offloading, 
AIGC service management, and incentive mechanism.

\subsection{Computation Offloading}

\begin{figure}
    \centering
    \includegraphics[width=\linewidth]{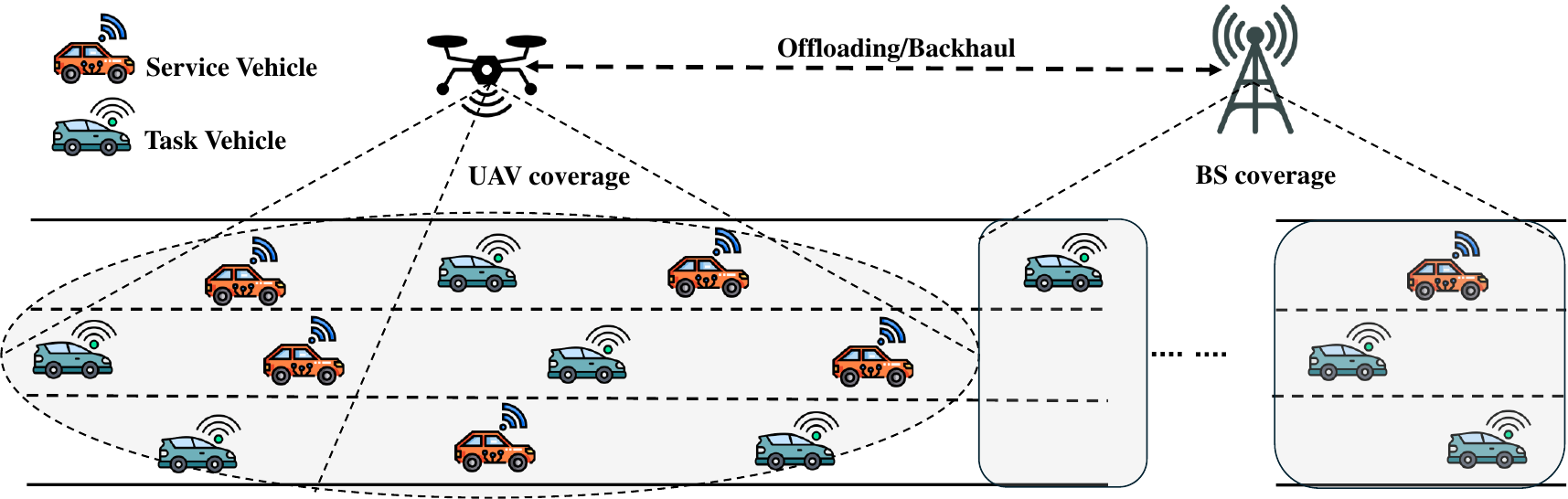}
    \caption{An unmanned aerial vehicle (UAV)-assisted highway connected autonomous vehicle system~\cite{cheng2024dependency}.}
    \label{fig:uav-cav}
\end{figure}

The studies in~\cite{cheng2024dependency,wu2024diffusion,liang2024diffsg} propose 
various DM-based approaches to optimize the user's offloading decisions in EC systems. 
For instance, the authors in~\cite{cheng2024dependency} investigate an unmanned aerial vehicle (UAV)-assisted highway connected autonomous vehicle system as illustrated in Fig.~\ref{fig:uav-cav}, where a task vehicle (TV) can offload its computing tasks to the nearby service vehicles (SVs). Each task can be divided into several dependent subtasks. 
The SV can also offload its received subtasks to a base station (BS) server via a relay UAV.
The offloading problem is formulated as a Markov decision process (MDP) which considers the subtask workload and interdependency information, available computing resources of SVs and BS, and distance from TV to SVs 
as a system state to select an offloading action. The reward function is to minimize the task completion delay between any two adjacent subtasks. 
To learn the optimal policy of the formulated MDP, the proposed diffusion-based DRL adopts a representative DRL algorithm, 
called synthetic experience replay (SER)-double deep Q-network (DDQN). 
Specifically, a generative diffusion model (GDM) is employed to generate the synthetic transition samples of the SER buffer 
which is used to train the DRL agent with improved convergence speed and sample efficiency. 
The simulation results show that proposed diffusion-based DRL approach can achieve about 38\% training reward improvement 
and 1.75x subtask completion time reduction, compared with the original DDQN algorithm.

Similarly, the study in~\cite{wu2024diffusion} considers an UAV-enabled EC network to provide ubiquitous Metaverse services for the distributed users which are equipped with the head-mounted (HMDs) to display 3D frames.
To achieve the low-latency Metaverse services, the users can offload their rendering tasks 
to a stationary EC and a flying UAV simultaneously via the dual connectivity communication channels. 
The rendering offloading problem is also modeled as a MDP 
in which the data size of rendering task, user-UAV and user-ES channel gains can be considered as a state 
to decide an offloading action on selecting the local HMD, ES or UAV for executing the rendering task. 
The objective is to minimize a weighted reward function of the rendering latency and total energy usage of 
the HDM and UAV. A diffusion probabilistic model (DPM)-based Metaverse rendering approach 
is proposed to generate the optimal offloading actions through the denoising and data-generation procedure. 
The simulations based on real-world datasets show that the proposed approach can reduce the rendering latency by 21.7\% and 54.7\%, compared the DRL-based offloading and random rendering algorithms, respectively.


Differently, the studies in \cite{tang2024dnn,huang2024adaptive,he2025qoi} develop various DM-based approaches 
for efficiently allocating the computing resources of ESs to execute the computing tasks offloaded from the users. 
For example, the authors in~\cite{tang2024dnn} investigate a networked UAV system in which each UAV acts as 
an ES to execute the deep neural network (DNN)-based target inspection tasks. 
The resource allocation problem is formulated as an MDP which considers the target positions, data size, DNN model type, latency requirement, and UAVs' energy capability and position as a state to select an action on splitting 
the DNN into multiple sub-models and assigning them for a set of specific UAVs. 
The objective is to maximize the reward relevant to the load balancing and task completion rates. 
A diffusion-based multi-agent DRL approach, called GDM-MADDPD which integrates a GDM into a multi-agent deep deterministic policy gradient (MADDPD) framework to learn the optimal assignment policies for multiple tasks at the same time. 
Specifically, the proposed GDM-MADDPD replaces the actor network of MADDPG with the denoising process of GDM 
for better capturing the correlation between the state and action during the training.
It can increase the task completion rate by 35\%, compared with the original MADDPD algorithm.

The study in~\cite{huang2024adaptive} proposes an adaptive digital twin (DT)-assisted communication, computing, and buffer control (3C) management framework for improving the quality of experience (QoE) in the multicast short video streaming systems. The adaptive 3C management problem is modeled as a MDP in which the user dynamics, channel gains, and DT model characteristic are considered as a state to select an optimal action on communication bandwidth and computing resource allocation as well as video segmentation version selection for the multicast groups in each resource allocation window. 
The objective is to maximize the long-term MG's QoE which is designed as a weighted function of multicast buffer time, video quality and quality variation. A diffusion-based twin delayed deep deterministic policy gradient (TD3) approach is proposed to learn the optimal 3C management policy employing the denoising process of a conditional diffusion model to help the actor network of TD3 explore the action space more effectively, thus generating robust actions to the network dynamics. 
As a result, the proposed approach can achieve about 18.4\% QoE improvement, compared with the four existing 3G management approaches.


Furthermore, the studies in~\cite{liang2024gdsg,liu2024dnn,du2024integrated} adopt the DMs for solving the joint offloading and resource allocation problem. For instance, the authors in~\cite{liang2024gdsg} formulate a mixed-integer NLP problem to jointly optimize the offloading decision and resource allocation for the users and ESs, respectively, with the objective of minimizing the weighted cost of delay and energy consumption. 
Then, a graph diffusion-based solution generator (GDSG) is proposed to find the optimal solution. 
Specifically, GDSG adopts a GDM which includes a convolutional graph neural network (GNN) to capture the distribution of high-quality solutions. Moreover, it employs the heuristic algorithms to efficiently create the high-quality suboptimal data samples to train the GNN. As a result, the GDSG can reduce the total task execution delay and energy cost by up to 42.07\%, compared with the existing learning-based and diffusion-based multi-task optimization solvers.

The authors in~\cite{liu2024dnn} investigate an AI-powered vehicular network in which 
a vehicle can offload a portion of its DNN to the ESs (i.e., nearby vehicles and road side units). 
A long-term mixed-integer NLP problem is formulated to jointly optimize the ES's resource allocation, 
vehicle's offloading and DNN partition decisions with the objective of minimizing the DNN task completion time, 
subject to the constraints on the available resources of ECs.
Then, the formulated optimization problem is transformed into sequential per-slot deterministic problems 
which are solved using a multi-agent diffusion-based deep reinforcement learning (MAD2RL) algorithm.
Specifically, MAD2RL adopts a multi-agent DRL learning framework, called QMIX~\cite{rashid2020monotonic}, 
including the DDPMs as the agent networks whose outputs are fed into a feed-forward mixing network
to evaluate the integrated decisions of all agents. 
Extensive simulations show that the MAD2RL can achieve about 52\% training reward improvement,
2.7x task completion time reduction, compared with the original QMIX~\cite{rashid2020monotonic}, genetic 
and greedy algorithms.


\subsection{AIGC Service Management}

\begin{figure}
    \centering
    \includegraphics[width=0.9\linewidth]{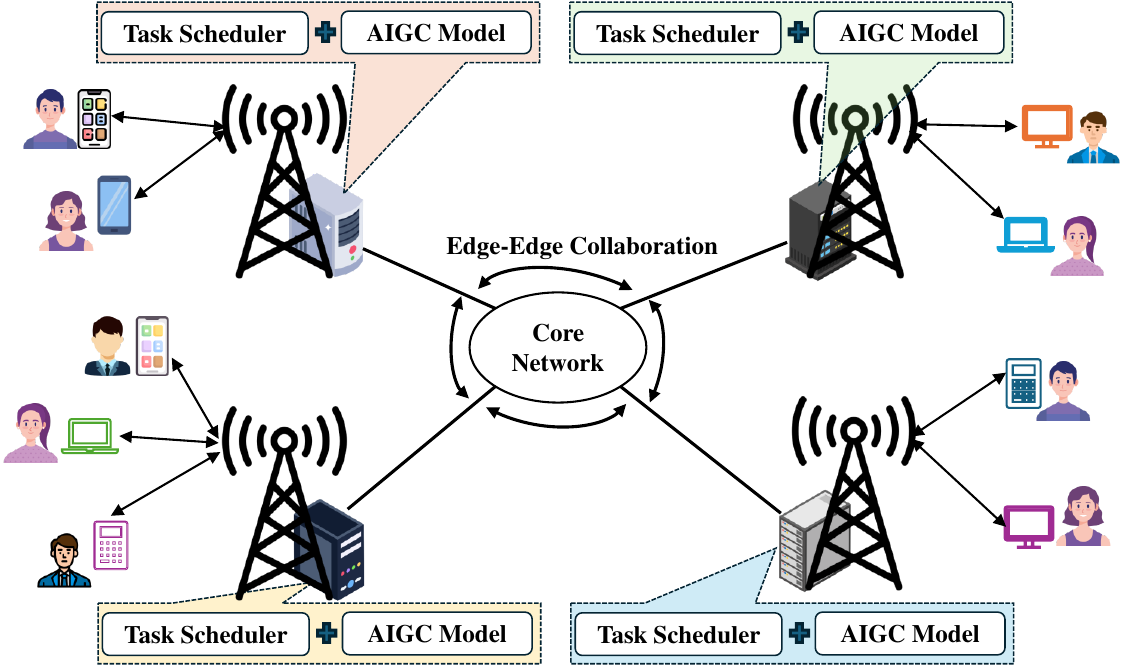}
    \caption{A distributed edge system with artificial intelligence generated content (AIGC) 
    services as proposed by~\cite{xu2024accelerating}.}
    \label{fig:edge-aigc}
\end{figure}

The above studies in~\cite{cheng2024dependency,wu2024diffusion,liang2024diffsg,tang2024dnn,huang2024adaptive,he2025qoi,liu2024dnn,liang2024gdsg,du2024integrated} consider general tasks whose computation workload is proportional to their data size. 
Differently, the studies in~\cite{xu2024accelerating,liu2025qos,du2023diffusion,akram2024ai} focus on 
managing the EC resources for the AI-generated content (AIGC) tasks whose workload is determined by the quality of required contents (e.g., generated images and videos). For instance, the study in~\cite{xu2024accelerating} investigates 
an AIGC task scheduling problem in a EC system where multiple ESs cooperate with each other to execute the DMs 
for generating the images as an AIGC service to the users as illustrated in Fig.~\ref{fig:edge-aigc}. 
Specifically, each ES has a task scheduler which decides to store its received AIGC tasks in its processing queue for 
local execution or offload the tasks to another ES via a wired core network. The task scheduling problem is modeled as a MDP which considers the workload and required computing resource of the new task as well as the total computation resource required to process all pending tasks of each ES as a state to select the ES for executing the new task. 
The reward function is designed to minimize the end-to-end delay of the new task. 
To learn the optimal policy, a diffusion-based DRL approach, called latent action diffusion-based task scheduling (LAD-TS) is proposed to leverage the laten action diffusion networks for balancing the trade-off between exploration and exploitation during the training. As a result, the proposed LAD-TS can reduce the required training time by at least 60\% and 
the image generation delay by 8.58\% to 33.67\%, compared with the baselines based on the DQN and SAC algorithms.  

The studies~\cite{liu2025qos,du2023diffusion,akram2024ai} aims at selecting an appropriate ES as the AIGC service provider (ASP) for each user request among the available servers equipped with one or multiple generative AI (GAI) models which can generate different types of AIGC. Specifically, they formulate the ASP selection problems in various AICG service systems 
as the MDPs that take the ESs' available computing resource, maximum completion time, and task computing resource requirement as a state to select an ASP for the AIGC task request. 
Then, they propose the diffusion-based DRL approaches which adopt a similar idea of using a specific DM to replace the multi-layer perception (MLP)-based policy network of the SAC algorithm for better capturing the complex relationships between the tasks and the ES attributes during the DRL training process. 
For example, in~\cite{liu2025qos}, the formulated MDP aims at selecting an optimal ASP for an user's image generation with the objective of maximizing the immediate reward as a weighted sum of the user's utility and penalty for the server crashes due to overload. Then, diffusion-based DRL approach is proposed to learn the optimal ASP selection policy by utilizing the reverse process of an attention-based diffusion model as the policy network of the SAC algorithm. 
As a result, it can improve the training and testing rewards by up to 37.5\% and 39.8\%, respectively, while reducing the server crash rate by up to 7.9\%, compared with the original SAC, Rainbow, PPO, and deep recurrent Q-Learning (DRQN) algorithms.

Similarly, the study in~\cite{du2023diffusion} formulates an MDP with a reward function which aims at maximizing 
the content quality of the AIGC output while minimizing the penalty due to the server overload. 
A GDM is adopted as the policy network of the SAC framework which can improve the training and testing rewards up to 1.4x 
and 1.2x, respectively, compared the representative DRL algorithms including the DQN, prioritized-DQN, REINFROCE, PPO, 
and SAC.
Similar to~\cite{du2023diffusion}, the authors in~\cite{akram2024ai} also integrate a GDM into the SAC algorithm 
to select the ASPs for executing the AIGC tasks which are required to create human digital twins (HDTs) for providing personalized healthcare functions in internet of medical things (IoMT)-based smart homes. 
The reward function is designed to maximize the quality of the generated contents 
while minimizing the penalties caused by the server overload and HDT task failure. 
As a result, the proposed approach can increase the task completion rate and overall system utility by up to 20\% and 15\%, respectively, compared with the seven baselines based on the heuristic and DRL algorithms.


\subsection{Incentive Mechanism}

The authors in~\cite{wen2024diffusion,liu2025contract,du2023ai} leverage the DMs for designing various incentive mechanisms 
which motivate both the users and ESs to join the EC systems. 
For instance, the study in~\cite{wen2024diffusion} considers an 6G-IoT network consisting of multiple ESs as 
the APSs equipped with the GAI models which are classified into different types according to their complexity. 
A user-centric incentive framework based on the contract theory is designed to motivate the ASPs to provide high-quality AIGC services to the user clients with the following three steps. In the first step, 
the client generates an incentive-compatible contract which specifies its service latency requirement 
and reward to the ASPs. In the second step, each ASP selects the best client contract according their model type and 
then takes the client-provided prompts as inputs to fine-tune their models using the real-time data collected from the IoT devices and generates the desired content. Upon receiving the generated contents, 
the client gives a reward specified in the contract to the ASP in the third step. 
In the proposed framework, a GDM is adopted as a contract generation network which maps the environmental state 
to an optimal contract which can maximize the expected cumulative utility of the client. 
Specifically, the Prospect theory is utilized to design the client utility which can effectively capture the risk 
attitudes of the client due to the lack of information on the GAI model types of ASPs. 
Then, the DQN framework is employed to train the GDM-based contract generation network, 
in which the optimal contracts are generated through the iterative noising and denoising processes. 
As a result, the proposed approach can improve the client utility up to about 3.48x, 
compared with the baseline based on the SAC algorithm.

The authors in~\cite{liu2025contract} introduce an edge Metaverse image generation framework 
in which the users subscribe with an image generation server located in the cloud. 
The user initiates its image generation request by uploading its captured images and text prompt to a mobile ES 
which then extracts the semantic data from the images and transfer them to the generation server for 
generating the high-quality images using the GDMs. The data transfer between the mobile ES and 
the generation server is critical to the overall performance of the image generation service. 
Thus, a contract-inspired contest theory-based incentive mechanism is proposed to design the payment plan 
for specifying a reward that the generation server should pay the ES according to the quality of the generated image influenced by the semantic information. 
Given the received reward, the ES hosts a contest game in which all semantic transfer tasks are incentive to 
choose a suitable transmit power level to earn a reward based on the semantic quality transfer. 
The payment design problem is formulated as an optimization problem to find the optimal user subscription fee, ES's reward and unit fee per quality of generation image which can maximize the image generation quality while ensuring efficient allocation of the ES's resources. Similar to~\cite{wen2024diffusion}, a GDM-based DRL approach is also proposed to iteratively refine the action space and reward structures during the DRL training process, which leads to the faster convergence 
due to the broader exploration ability. As a result, the proposed approach can reduce the training time by up to 1.68x while improving the reward after convergence by up to 28.39\%, compared with three baselines based on the PPO, SAC, and transformer-based SAC algorithms.

Moreover, the authors in~\cite{du2023ai} design a full-duplex device-to-device (D2D) semantic communication framework 
for efficient information sharing in a mixed reality (MR)-aided Metaverse system. 
The nearby MR users equipped with HMDs can share their semantic information extracted from their view image with each other. Then, a user leverages its received information to efficiently render the view images displayed on its HMD through lightweight semantic matching. As such, the user's computing resource usage can be significantly reduced. 
To motivate the users to participate in this information sharing framework, a contract theory-based incentive mechanism 
is proposed to design the contract which specifies the payment made by the semantic information receiver (SIR) to the semantic information provider (SIP) and the fee charged for a unit quality of shared semantic information (QoS) value. 
A conditional diffusion model is adopted to generate the contract design policy by mapping an environment state to an optimal contract which can maximize the utility of the SIR and provide the SIP with the necessary incentive to agree on the contract.
The numerical results show that the proposed incentive mechanism can achieve a faster learning convergence speed with an SIR utility improvement of up to 1.02x, compared with the incentive mechanism baselines based the PPO and SAC algorithms.





\subsection{Lessons Learned}

From the above review, DMs have been widely integrated with the DRL algorithms for optimizing 
the resource management in the EC networks. Specifically, the DMs allow the DRL agents to sample the diverse actions or state-action trajectories through the iterative forward and reverse processes. 
Thus, the diffusion-based DRL agents can explore the dynamic environments of the EC networks more effectively. 
Moreover, the DM's ability of representing the complex action distributions also helps the DRL agents 
avoid getting stuck in suboptimal local policies during the training. 
The DMs can also be used to generate the synthetic high-quality state-action trajectories samples in the offline DRL training approaches. As a result, the DMs can improve the sample efficiency and learning policy of the DRL-based edge resource management approaches. 

However, integration of the DMs also increases the computation overhead of the DRL algorithms due to involvement 
of the compute-expensive denoising steps. Thus, the diffusion-based DRL approaches may not be applied for making the real-time resource management decisions in the latency-critical EC networks. 
Moreover, training and running such compute-intensive diffusion-based DRL approaches may not be suited
for the user devices which are typically equipped with the limited computing resources. 
This calls for new research to reduce the computational overhead of the diffusion-based DRL algorithms. 
A promising approach is to adopt the compression techniques such as the knowledge distillation, pruning, quantization, and fine-tuning to transform the compute-intensive DMs into the lightweight DMs~\cite{song2024lightweight}. 
Such lightweight DMs can be integrated into the DRL frameworks for making the real-time actions in the EC networks.

\section{DMs for Semantic Communication}
\label{Sec:SemCom}

\begin{figure}
    \centering
    \includegraphics[width=0.9\linewidth]{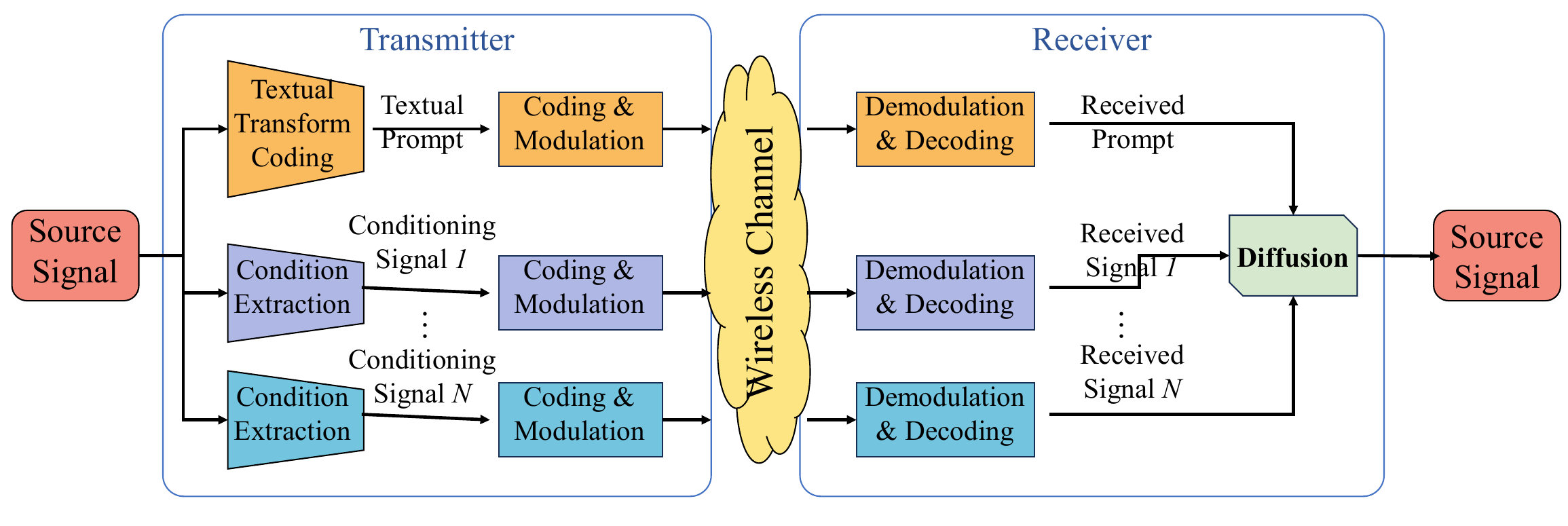}
    \caption{System architecture of the proposed semantic communication framework empowered by DMs \cite{qiao2024latency}. The transmitter encodes the source signal into a textual prompt and multiple conditioning signals, which are transmitted over the wireless channel. At the receiver, the prompt and conditioning signals are jointly used to reconstruct the original signal via a DM process.}
    \label{fig:SemCom1_1}
\end{figure}

As shown in Fig.~\ref{fig:SemCom1_1}, DMs have emerged as powerful generative techniques that significantly enhance semantic communication systems, providing effective solutions to traditional challenges such as noise robustness, joint optimization of source-channel coding, and multi-modal semantic integration. This section comprehensively reviews recent advancements leveraging DMs in semantic communication, highlighting their substantial contributions across various key application scenarios and research directions.

\subsection{DMs for Joint Source-Channel Coding and Semantic Reconstruction}

Semantic communication faces significant challenges from channel-induced noise, which can severely degrade the quality and accuracy of transmitted semantic information. Recently, DMs have emerged as promising solutions to address these challenges by leveraging their intrinsic denoising capabilities and robust generative mechanisms.

Several recent studies have specifically explored diffusion-driven frameworks for handling general and specific channel noise scenarios. For instance, the authors in~\cite{guo2024diffusion} proposed a diffusion-driven semantic communication framework explicitly tailored to handle general channel noise conditions. They creatively mapped wireless channel noise characteristics into the $T$-th step of the forward diffusion process, allowing the receiver to effectively perform progressive denoising and signal reconstruction via the reverse diffusion mechanism. Their framework also incorporated downsampling and upsampling modules, reducing bandwidth consumption while ensuring that reconstructed features adhered to Gaussian distributions via a variational autoencoder (VAE). However, their approach introduced additional complexity due to the precise tuning required for the diffusion steps.

\begin{figure}
    \centering
    \includegraphics[width=0.9\linewidth]{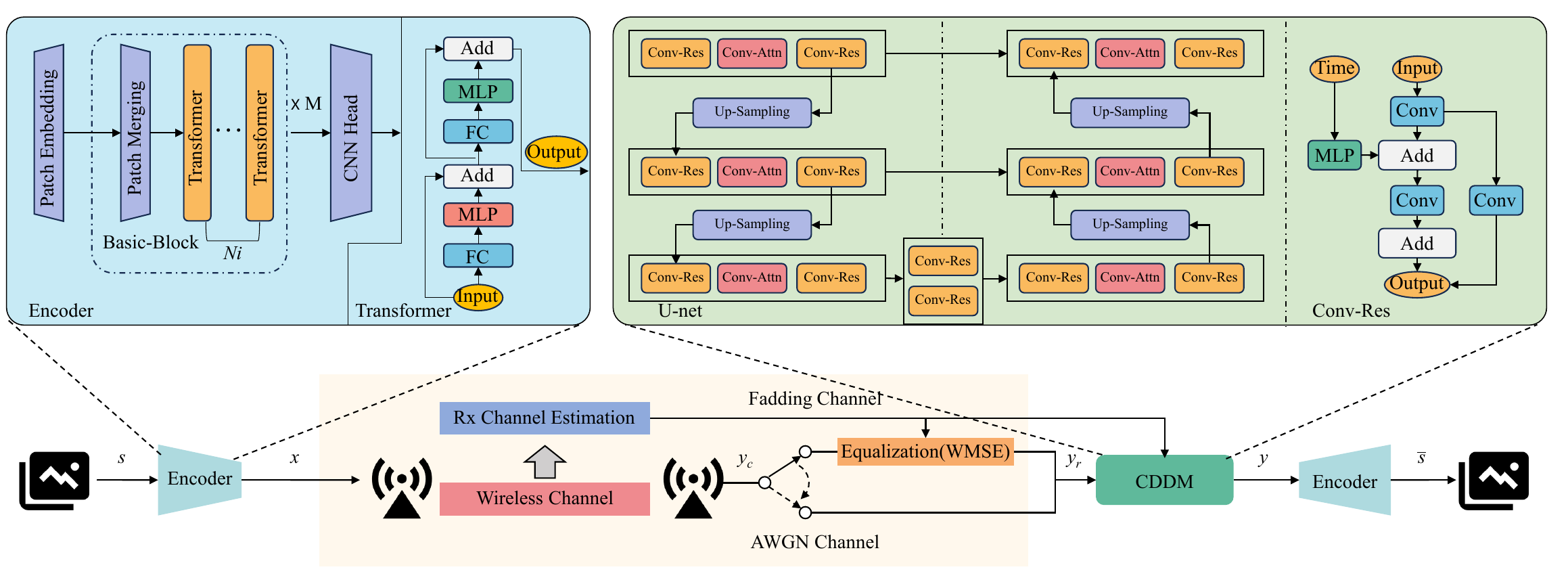}
    \caption{Overall architecture of the proposed semantic communication system with DM \cite{wu2024cddm}. The source image is encoded and transmitted over a fading wireless channel with channel estimation and WMSE-based equalization. At the receiver side, a CDDM reconstructs the image using a U-Net backbone with multi-resolution attention and residual modules. Transformer-based components are integrated to refine semantic features and facilitate accurate recovery.}
    \label{fig:SemCom1_2}
\end{figure}

Addressing the specific complexity of Multiple-Input Multiple-Output (MIMO) channels, two closely related works proposed tailored diffusion-based solutions. The authors in~\cite{duan2024dm} designed a diffusion model leveraging singular value decomposition (SVD) to decompose complex MIMO channels into parallel sub-channels. They developed a joint adaptive sampling algorithm that dynamically assigned diffusion sampling steps based on estimated effective noise power, effectively enhancing reconstruction robustness under dynamic MIMO channel conditions. Similarly, as shown in Fig.~\ref{fig:SemCom1_2}, the authors in~\cite{wu2024cddm} proposed a Channel Denoising Diffusion Module (CDDM), strategically placed after channel equalization to explicitly model specific wireless channel noise patterns. Unlike Duan et al., who focused primarily on adaptive sampling, CDDM employed a custom forward and reverse diffusion process specifically designed around noise characteristics, resulting in substantial reconstruction accuracy improvements, such as approximately 1.06 dB reduction in Mean Square Error (MSE) in terms of Peak Signal-to-Noise Ratio (PSNR) under additive white Gaussian noise (AWGN) and Rayleigh fading channels. Nonetheless, both these MIMO-focused methods required extensive training tailored to individual channel scenarios, increasing their computational complexity and deployment overhead.

Moving beyond channel-specific approaches, the authors in~\cite{zhang2025semantics} introduced semantic guidance directly into the diffusion denoising mechanism. In contrast to traditional denoising approaches that only considered physical channel conditions, their method incorporated semantic cues—such as textual descriptions and edge maps—as conditional inputs during the denoising process. This semantic-driven diffusion model effectively preserved semantic integrity, demonstrating significant robustness even under severe channel disturbances and uncertain channel state information (CSI) conditions at both transmitter and receiver. Nevertheless, while semantic guidance improved semantic fidelity substantially, it imposed extra computational burdens due to semantic feature extraction and integration processes.

In low-SNR scenarios, the authors in~\cite{nguyen2024leveraging} further exploited Stable DMs combined with context-aware semantic prompts, which were generated using Contrastive Language-Image Pre-training (CLIP). Their approach uniquely provided crucial semantic contexts, significantly stabilizing semantic reconstruction quality even under extreme channel noise. Empirical results validated on the Kodak dataset showed superior image fidelity and perceptual quality compared to traditional methods, particularly highlighting its robustness at very low SNR levels. However, generating and processing these additional semantic prompts incurred higher computational resource demands, posing potential limitations for deployment in resource-constrained settings.


\subsection{DMs for Multimodal and Cross-Modal Semantic Communication}

Multimodal and cross-modal semantic communications have recently attracted substantial research interest due to their ability to exploit semantic correlations across different modalities, significantly enhancing information transmission efficiency and robustness. 

Several recent studies have specifically explored multimodal semantic communication leveraging diffusion-based text-to-image generation frameworks. For instance, the authors in~\cite{wei2024language} developed a framework where input images were initially converted into textual modality data at the transmitter side and then transmitted through noisy channels. At the receiver, a DM reconstructed the original images directly from textual descriptions, significantly enhancing transmission robustness. Similarly, the authors in~\cite{liang2025vision} further advanced this idea by utilizing Stable Diffusion integrated with vision-language models (VLMs) for massive MIMO semantic communication. Unlike Wei et al.'s general-purpose model, Liang et al. explicitly focused on reducing bandwidth by transmitting semantic textual descriptions of images rather than raw image data. Although their approach yielded moderate results in terms of SSIM and PSNR metrics, it achieved superior semantic integrity, effectively prioritizing semantic accuracy over pixel-level fidelity. However, both approaches introduced extra computational complexity.

Beyond the text-image modality, the authors in~\cite{fu2024multimodal} proposed a multimodal generative semantic communication system specifically designed for visible-light and infrared modalities, particularly suitable for emergency scenarios. Their transmitter extracted and compressed semantic segmentation maps, employing one-hot encoding and zlib compression to enhance robustness. The receiver utilized a latent DM combined with contrastive learning to simultaneously reconstruct visible RGB and infrared images. This innovative multimodal integration framework demonstrated notable performance gains, achieving a 3\% classification accuracy improvement and a 5.5-point reduction in FID score over existing methods. Similarly targeting multimodal feature integration, the authors in~\cite{liang2024image} introduced a Multimodal Semantic Communication (MMSemCom) framework that utilized both image features and textual prompts, extracted via Contrastive Language-Image Pre-training (CLIP), to guide conditional image reconstructions at the receiver. Compared to Fu et al.'s modality-specific approach, MMSemCom offered a more generalizable multimodal reconstruction capability, significantly enhancing performance with PSNR gains of 5.95 dB and 5.85 dB on CIFAR-100 and STL-100 datasets, respectively.

Focusing on more sophisticated multimodal tasks, the authors in~\cite{yuan2024generative} investigated a semantic communication system that simultaneously performed image reconstruction and segmentation. Their design incorporated semantic knowledge bases (KBs) at both transmitter and receiver, leveraging Swin Transformers and ResNets to guide joint source-channel decoding. Moreover, task-specific KBs facilitated adaptive semantic feature selection based on predefined instructions. This sophisticated integration of semantic knowledge and diffusion-based reconstruction significantly improved both PSNR for image reconstruction and Intersection-over-Union (IoU) for segmentation tasks, outperforming conventional techniques such as Deep JSCC and JPEG+LDPC+QAM. However, the complexity of maintaining multiple KBs posed considerable computational and storage overhead.

Further extending multimodal semantic communications into immersive media scenarios, the authors in~\cite{zhang2024diffusion} presented a novel diffusion-based semantic communication framework explicitly designed for Virtual Reality (VR) dual-fisheye images. This pioneering framework simultaneously performed semantic extraction, transmission, and panoramic stitching. Leveraging a multi-scale semantic condition extractor with self-attention mechanisms, the DM could effectively distinguish meaningful VR content from peripheral artifacts, providing high-quality panoramic reconstructions. Compared to Yuan et al.'s general image-semantic integration, Zhang et al. provided an explicitly tailored framework for immersive VR scenarios, significantly advancing multimedia communication capabilities. Nevertheless, the real-time constraints of VR applications raised practical challenges regarding inference speed and system latency.

Finally, targeting next-generation vehicular networks, the authors in~\cite{lu2024generative} proposed a Generative AI-Enhanced Multi-Modal Semantic Communication (G-MSC) framework specifically for the Internet of Vehicles (IoV). Their approach combined bird's-eye-view (BEV) fusion for efficient multimodal integration, GAN-based channel estimation for robustness, and DMs for semantic inference and prediction. Empirical validations on the nuScenes-mini dataset showed substantial transmission overhead reduction (40-60\%), while maintaining high semantic fidelity. Unlike previous modality-specific or general multimodal methods~\cite{fu2024multimodal, liang2024image}, G-MSC explicitly focused on dynamic, mobility-intensive scenarios in vehicular contexts, achieving a more practical balance between computational complexity and real-time semantic integrity.

\subsection{DMs for Semantic Communication Efficiency and Resource Constraints}

Efficiency and resource management are critical challenges in semantic communication systems, especially under constraints such as limited bandwidth, computational resources, and energy budgets. DMs have recently emerged as effective solutions addressing these constraints, providing robust semantic transmission with significantly improved efficiency.

Several recent works have specifically targeted the deployment of DMs in environments with stringent computational and resource constraints. For instance, the authors in~\cite{pignata2024lightweight} introduced a quantized generative semantic communication system tailored explicitly for resource-limited scenarios. They leveraged a quantized latent DDPM at the receiver, employing post-training quantization (PTQ) to compress diffusion model parameters effectively. By utilizing adaptive rounding and calibration techniques, their approach significantly reduced both computational load and memory usage, thereby validating its suitability for deployment on resource-constrained edge devices. However, this quantization approach could potentially lead to accuracy degradation in more complex semantic scenarios, necessitating careful trade-offs between efficiency and quality.

Extending the notion of resource-aware semantic communication, the authors in~\cite{li2024goal} and~\cite{xu2023latent} independently developed frameworks focused on conditional and adaptive denoising under resource constraints. Specifically, the authors in~\cite{li2024goal} proposed a goal-oriented semantic communication system, termed SD-GSC, leveraging Score-based Stochastic Differential Equations (SSDE). Their semantic encoder and conditional semantic denoiser effectively utilized instantaneous channel gains as inputs, significantly enhancing semantic reconstruction quality. Empirical evaluations showed approximately 32\% improvement in PSNR and a 40\% reduction in FID compared to state-of-the-art approaches. Similarly emphasizing denoising under stringent conditions, the authors in~\cite{xu2023latent} proposed the Latent Diffusion-based De-Noising Semantic Communication (Latent-Diff DNSC) system for robust semantic inference. This framework integrated a Variational Autoencoder (VAE), adversarial learning, and DMs, achieving remarkable gains—up to 5.7 dB in PSNR improvement compared to ADJSCC methods and surpassing DeepJSCC by 20–67\% in PSNR and 4–68\% in SSIM metrics. Although both approaches demonstrated excellent semantic reconstruction under challenging channel conditions, they incurred additional complexity due to intricate conditional mechanisms and adversarial components, potentially impacting their applicability in highly latency-sensitive scenarios.

Another direction focused explicitly on latency reduction. The authors in~\cite{wang2024temporal} and~\cite{wang2024fast} independently explored temporal prompt engineering methods. The former proposed a Temporal Prompt Engineering-based Generative Semantic Communication (TPE-PGSC) framework, featuring parallel semantic extraction and generation mechanisms optimized by reinforcement learning, which significantly reduced latency by 52\% with a minor 9\% reduction in semantic accuracy. In a related but distinct work, the authors in~\cite{wang2024fast} developed a generative semantic communication mechanism employing sequential conditional denoising guided by reinforcement learning-driven temporal prompt engineering. This method dynamically adapted semantic transmission sequences, maintaining similar accuracy to traditional methods but achieving a notable 52\% latency reduction. Both methods underscored the effectiveness of integrating temporal considerations into diffusion-based semantic communication, but the complex prompt engineering procedures added computational overhead that may limit deployment in extremely constrained environments.

Focusing specifically on ultra-low-bit-rate scenarios, the authors in~\cite{mao2024diffcp} introduced DiffCP, a collaborative perception framework utilizing geometric priors and semantic cues to reconstruct Bird’s Eye View (BEV) features from highly compressed semantic transmissions. DiffCP required only 80 Kbps bandwidth, significantly reducing communication overhead by a remarkable 14.5-fold compared to existing algorithms, demonstrating its strong suitability for intelligent transportation systems. In contrast, the authors in~\cite{zhang2024semantic} presented Semantic Successive Refinement, a multi-stage transmission strategy employing diffusion-based conditional vector estimation. Their approach dynamically improved image quality by adaptively leveraging hierarchical feature extraction and caching strategies as additional bandwidth became available. Both studies demonstrated outstanding bandwidth efficiency, yet Mao et al. focused on a static ultra-low-rate scenario, while Zhang et al. targeted dynamic, adaptive conditions requiring more complex caching and hierarchical structures.

Exploring satellite communication (SatCom) environments, the authors in~\cite{jiang2024semantic} integrated SegGPT with conditional DMs to design a robust semantic communication system. Their framework adaptively managed dynamic satellite channel conditions, substantially outperforming conventional methods such as JSCC and JPEG combined with LDPC coding. Despite superior performance gains, their approach's complexity, particularly in integrating SegGPT and DMs, raised potential implementation challenges for resource-constrained satellite deployments.

Targeting semantic communication in IoT environments explicitly, the authors in~\cite{ren2023asymmetric} proposed an asymmetric semantic communication framework leveraging lightweight DDPM and ResNet-based encoders. Their system provided adaptive semantic reconstruction through re-training on IoT devices with varying channel conditions, significantly enhancing robustness against interference. Nonetheless, the method still suffered from slow inference speeds due to lightweight model constraints, indicating the necessity for further computational optimization in highly constrained IoT scenarios.

Lastly, the authors in.~\cite{yang2024semantic} and~\cite{yang2024agent} independently developed semantic-driven adaptive frameworks focusing on resource efficiency in remote monitoring and surveillance tasks. the authors in~\cite{yang2024semantic} defined a new semantic metric termed Value of Information (VoI), integrating Age of Information (AoI) with semantic change detection, effectively optimizing bandwidth usage and semantic quality. Concurrently, the authors i~\cite{yang2024agent} proposed an Agent-driven Generative Semantic Communication (A-GSC) framework combining generative AI and reinforcement learning, achieving substantial energy reductions (40–60\%) compared to periodic sampling. While both methods successfully optimized semantic resource efficiency through adaptive sampling and monitoring mechanisms, they introduced additional computational overhead from agent-based learning and dynamic semantic evaluations.

\subsection{Lesson Learned}
The reviewed studies underscore the transformative potential of DMs in addressing critical challenges in semantic communication. Specifically, diffusion-based approaches have demonstrated significant advantages in enhancing noise robustness, optimizing joint source-channel coding for efficient semantic reconstruction, and effectively facilitating multimodal semantic interactions. Furthermore, their inherent adaptability to resource-constrained environments makes them exceptionally suitable for practical deployment. Nonetheless, future research directions should focus on refining computational efficiency, reducing inference latency, and improving scalability to fully exploit the generative capabilities of DMs in complex, real-world semantic communication scenarios.


\section{DMs for Other Issues}
\label{sec:DM_for_other_issues}


\subsection{Wireless Security}


Wireless security has been always a key research issue in wireless communications. 
DMs with denoising capabilities can enhance resilience of wireless communications systems. In addition, DMs excel at data generation capabilities, and thus generated synthetic data can be used to train and improve security algorithms.  
For example, the work in \cite{su2024hybrid} 
develops a new framework of multi-modal LLM to address the challenges of data freshness, security, and incentivization in the Internet of Medical Things. Unlike conventional uni-modal LLM, this work develops a multi-modal RAG-empowered LLM solution as data is normally created and stored in decentralized data locations (e.g., hospitals) and as the integration of medical data from multiple sources can improve the performance and functionality of healthcare systems. In addition, this work leverages the concept of blockchain to facilitate secure data management in healthcare systems and avoid the single point of failure issue due to centralized data management. Experimental results show that the proposed method with DM-based contract theory and blockchain for data sharing can significantly improve performance over conventional learning approaches, e.g., around $20.4\%$ compared with DRL-PPO.

DMs have demonstrated their potential for improving semantic communications, as reviewed in Section~\ref{Sec:SemCom}. In this area, several studies (e.g., \cite{ren2023asymmetric, ren2024diffusion, zheng2024energy, du2024generative}) have also leveraged DMs as a suitable tool to improve semantic communication security. For example, the work in \cite{ren2023asymmetric, ren2024diffusion} addresses vulnerability issues in semantic communications by exploiting DM and DRL. In addition to the basic components of a semantic communication system, the proposed architecture adds a diffusing module at the sender and an asymmetric denoising module at the receiver.
While the diffusing module adds Gaussian noise to dominate adversarial perturbations (i.e., diffusion purification), the asymmetric denoising module is employed at the receiver side to address both semantic attacks at the source and channel noises over the communication channels, thereby enhancing the robustness (against semantic attacks) of the system. By simulation results, it is shown that the proposed method, named DiffuSec, can achieve a classification accuracy of up to $91.1\%$ and outperform a couple of benchmarks using a conventional vision transformer and vector quantization-variational autoencoder as well as conventional channel coding schemes using Low-Density Parity Check (LDPC).
%
%
%

Similarly, \cite{zheng2024energy} exploits the power of generative DMs to develop a more universally applicable solution for secure semantic communications. In addition, the work considers balancing defence efficiency and energy efficiency, thereby making the proposed method suitable for implementation in resource-constrained mobile devices.
The work in \cite{du2024generative} shows that GenAI and DM have the potential to improve the security of semantic communications. In particular, this work proposes to combine the structural fidelity of visual prompts and the semantic information of textual prompts (i.e., multi-model prompts) to enhance the stability of vanilla GenAI-assisted decoders. However, the transmission of visual or textual prompts in wireless environments may lead to severe security issues, for example, attackers can use visual prompts of an objective to find its original image. This challenge is addressed through the use of covert communications \cite{liu2025generative} by imposing the detection error probability constraint in the joint optimization problem of diffusion step, transmit power and jamming power allocation. The security challenge of such studies in wireless communications in general and semantic communications in particular can also be overcome by applying advanced 6G technologies, such as, intelligent reflecting surface (IRS) \cite{liu2024physical}. In such works, generative DMs can be utilized to enhance the action spaces of conventional RL solutions (e.g., DDPP and PPO) through DM-generated high-quality data samples.

Moreover, DMs have been investigated for the security improvement of space-air-ground integrated networks \cite{liang2024uav, yao2024biometric, zhang2024multi, he2025synthetic}, where the network access and wireless services can be provided by different network tiers, such as terrestrial communications, UAVs, HAPs, and satellite networks \cite{dao2021survey, pham2022aerial}.
In \cite{liang2024uav}, the optimization problem to minimize the secure age of information objective is investigated by optimizing user scheduling for IoT devices, trajectory of the UAV, and time allocation for the data transmission slots. To overcome the non-convexity of this problem, a solution is designed by integrating the TD3 approach with a DM. Via the capabilities of DM to enhance the action space, the proposed DM-TD3 solution can outperform several RL benchmarks, such as the original TD3 scheme, DDPP, and SAC, in terms of secure age of information and energy consumption. 
Similarly, the work in \cite{zhang2024multi} integrates generative DM and TD3 to develop a new solution, called GDMTD3, to solve the problem of maximizing the secrecy rate and UAV energy consumption in UAV swarm systems. 
%
%
Although promising, the training of DMs normally requires a large amount of data, which may contain sensitive information, and if leaked, it may cause severe privacy issues. Therefore, the work in \cite{yao2024biometric} leverages differential privacy to enhance privacy protection of the data used for training the DM model in UAV delivery networks. 
%
%
%
In \cite{he2025synthetic}, a four-layer federated learning (FL) approach is developed for integrated satellite-terrestrial networks. In particular, the network is composed of four layers, including 1) the local layer that contributes data for local model training in FL and edge aggregation, 2) the edge layer where each acts as a client in FL, 3) the satellite layer that acts as transponders, and 4) the global layer that updates the global model in FL based on model updates received from the edge layer via the satellite layer. Founding up on this network architecture, the work develops a new solution that combines a conditional DM to ensure global data representation and differential privacy to ensure data privacy protection for intrusion detection tasks. Using a real dataset, the proposed method is shown to have superior performance over a couple of FL approaches (e.g., FedAvg and FedProx \cite{MLSYS2020_1f5fe839}), for example, it can achieve the accuracy of $96.63\%$ and the precision of $96.71\%$ in the non-IID setting.

In addition, the security in the metaverse and sensor networks can also be addressed through the use of DMs. For example, \cite{kang2024hybrid} develops a new hybrid DRL for security in vehicular metaverse networks. Unlike other works on DM-based RL solutions, the hybrid solution proposed in \cite{kang2024hybrid} can generate both continuous actions (i.e., proportion of pre-migration tasks) and discrete actions (i.e., migration action). Via simulations, the proposed hybrid solution using generative DM can offer significant performance improvement over the baselines, such as $6.06\%$ compared to the baseline without pre-migration consideration and $62.52\%$ compared to the multi-agent PPO method.
\cite{zhao2025generative} models the optimization problem of sensor placement for anomaly detection in cyberphysical systems. This NP-hard problem can be solved via the proposed algorithm, namely experience feedback graph diffusion (EFGD) method, which leverages DMs to generate high-quality sensor placements and human feedback mechanisms to expedite convergence. This method shows notable performance improvement over advanced DM methods, including GDPO \cite{liu2025graph} and DDPO \cite{black2023training}, in a real power system.

\subsection{Radio Map Estimation}
Radio frequency map graphically represents the strengths of RF signals at different locations, which facilitates network operators to deploy network infrastructure, e.g., access points, to maximize network coverage and reliability. However, collecting a huge amount of data for the radio frequency map is costly. DM has demonstrated its effectiveness in data generation, and thus it can be applied for generating the radio map as proposed in \cite{luo2025denoising}. Accordingly, multiple receivers are deployed at different locations of an area of interest to collect the RF signals transmitted from a transmitter. The RF signal data is then used as a condition of the DM, which uses Latent DM \cite{rombach2022high} architecture. By learning the denoising process, the DM generates new data that resembles the training data. Simulation results in both indoor mmWave and outdoor sub-6GHz networks show that the proposed DM is capable of generating radio maps with precision up to $95$\%. It is noted that the generated radio map can be transmitted to other devices for further processing, resulting in communication cost due to the large size of the radio map. In this case, semantic communication can be used to extract and transmit important features of the radio map as presented in~\cite{zhou2025generative}.

\subsection{Spectrum Trading}
Spectrum trading between mobile users and a service provider can be modeled using a Stackelberg game, which traditionally requires full network information to find equilibrium. However, finding the equilibrium requires complete network information. A recent work~\cite{liu2024optimizing} demonstrates that DMs can effectively find this equilibrium by leveraging their reverse denoising process. In~\cite{liu2024optimizing}, multiple users request AIGC services by submitting prompts to a service provider. A DM-based semantic communication technique is implemented to extract the semantic information of the images for the users. The concept of age of semantic information dependent on allocated bandwidth is proposed to model semantic information freshness. Then, the Stackelberg
game problem is formulated where the service provider as a leader optimizes the bandwidth price and the users as followers maximize the number of bandwidth units. The DM is then used to find the game equilibrium. Simulation results show up to $80\%$ improvements in SSIM and PSNR over VAE-based methods, and the DM-enabled scheme achieves near-optimal payoffs despite incomplete information. However, this work only considers the single-provider scenario, and image reconstruction at the users may cause high computation cost and image quality distortion.

\subsection{User Association}
User-BS association is a prerequisite for the users to access the networks. However, designing an effective association scheme requires user trajectories and spatial data, raising acquisition cost and privacy issues. As DM is effective in generating unseen data, it can be used to model user mobility for user-BS association as presented in~\cite{tao2024parallel}. Particularly, the system model consists of multiple users and BSs. The problem aims to determine the user-BS association maximizing total data rate while minimizing handover cost. A PPO-based multi-agent DRL scheme is used in which each user locally selects the BS based on SINR, whose distribution is determined by the user trajectory models. A DDPM \cite{ho2020denoising} with attention-based U-Net~\cite{ronneberger2015unet} denoising is used to generate trajectories with street maps as conditioning. Simulations with the real dataset of vehicle trajectories and OpenStreetMap~\cite{StreetMat-data} show that the proposed DM can produce lifelike trajectories even in zero-shot settings. As a result, the DRL-DM scheme achieves up to $91.6$\% of the cell-edge user performance and $96.6$\% of the network utility compared to agents trained in the real environment.

\subsection{Access Control}

The authors in \cite{liu2024generative} investigate access control in an IEEE 802.11be WiFi system where the DM-based DDPG algorithm is used to optimize the contention window and the aggregation frame length for each user to maximize the total throughput. Specifically, the decision network in the DDPG is based on the reverse process of the DM, with channel idle time proportion and packet loss rate used as state information. Simulation results show that the DM-based DDPG improves the throughput up to $10.5$\% while reducing the average latency to $10$ ms compared with traditional DDPG. The reason may be that the use of DM enables better modeling of the relationships between the system states and actions. However, how the throughput of the users depends on the system states is not clearly discussed in this work.

The authors in \cite{li2025offline} consider a user scheduling problem in which a scheduler receives data packets and delivers them to users. The problem aims to determine amounts of resources allocated to the packet transmissions of the users to maximize the total throughput subject to the average resource consumption constraint. The offline learning challenge is considered, in which the solver cannot interact with the environment and can only deduce the optimal policy using a collected dataset. As a result, an offline RL-based scheduling algorithm is introduced with an SSDE for learning the policy from dataset and a critic model for approximating $Q$-functions. The actions, i.e., solutions, are sampled from the SSDE-generated distribution and the final action is taken with regard to the maximum confidence score calculated by applying softmax on the $Q$-values. Simulation results show an improvement of up to $8\%$ in throughput compared to actor-critic method.


\subsection{Power Control}
The work in \cite{wang2024generative_rl} investigates an uplink wireless RSMA network consisting of multiple LEO serving different ground terminals. The problem aims to optimize transmit power of the terminals and receive beamforming, i.e., receive filter, to maximize the total throughput. To solve this problem, DM-based PPO is used, with a DM as the policy-inducing actor network, which facilitates more efficient parameter tuning and sample efficiency. Simulation results show that the use of DM enables the PPO to improve the throughput up to $25.87$\% due to its ability to capture intricate patterns and relationships within the environment. Moreover, the use of RSMA helps to improve the sum rate up to $10$\% compared to NOMA. 

DM is also utilized in power control for trains in train-to-train (T2T) communication network as presented in~\cite{ye2024enhancing} or in an RSMA-enabled cell-free massive MIMO system as proposed in \cite{zheng2024rate}. Particularly, in~\cite{zheng2024rate}, the problem aims to optimize the power-splitting factors for the common message transmission of the BSs and the power to maximize the spectrum efficiency over the users. To obtain a generative solution under different network environments, the DM is used with power-splitting factors and power-control coefficients as input data. The objective of DM training is to minimize the difference between the generated spectrum efficiency and a target one obtained by a genetic algorithm. Simulation results show that compared to PPO-based DRL, the proposed DM can improve the spectrum efficiency up to $60$\%. However, a heuristic algorithm is required for guiding the optimization towards optimal solution, which is time-consuming. A simpler solution, i.e., iterative random shooting method~\cite{michalik2009incremental}, can be used to predict the outcomes of the action sequences as proposed in~\cite{ye2024enhancing}. This method is particularly suitable to highly dynamic and complex networks such as T2T communications. 

Different from \cite{zheng2024rate}, the authors in \cite{du2023yolo} consider a semantic communication-enabled DT system where a UAV takes images and uses the YOLOv7-X object detector to extract the object-level semantics. Each object is assigned an importance score based on the detector’s confidence, guiding transmission priority. Then, the problem aims to determine the importance score and transmit power associated with each object to maximize the total importance score and the SSIM. DM-enabled DRL is employed, where the environment consists of wireless channels and object count of the image. Power allocation policy is achieved through the denoising process that is trained by a double Q-learning method. Simulation results show an improvement of up to $1.1$\% in terms of total score of the proposed DM scheme compared to the average power transmission scheme. However, the proposed work does not discuss the DT performance. SSIM metric should also be used to show the efficiency of the proposed model.


\subsection{Data Collection}

In \cite{li2024diffusion}, the system model consists of UAVs that collect and transmit data to a remote cloud for the DT synchronization. The UAVs can select satellites as relays or directly transmit the data to the remote cloud via a BS. Given the UAVs' energy constraints, the problem is to make decisions on relay selection, transmit power, and transmit time allocation to the UAVs to minimize their total energy consumption. The transmit power and time allocation are determined by using traditional tools, i.e., CVX, while the relay selection problem is solved by the conditional DM. Therein, the UAVs' data arrival rates and fading channels are the state, and the gradient-based sampling~\cite{janner2022planning} is used to guide the relay decision sampling procedure. Simulation results show that the proposed DM can reduce the UAVs' energy consumption up to $35$\% and $22$\% compared with the genetic and traditional DRL algorithms, respectively. In addition, the proposed DM reduces the queue length at the UAVs. However, UAVs' trajectory optimization and the energy consumption for the UAVs propulsion are not considered in this work. To speed up the data collection, unmanned ground vehicles (UGVs) are used along with the UAVs for the data collection as proposed in \cite{zhao2024energy}. To enhance energy efficiency, a minimum number of stop points for data collection of the UGVs needs to be determined to guarantee the QoS of the collected data. For this, a DM-based SAC approach is proposed with uniform noise perturbation. 

The authors in \cite{zheng2025uav} consider a post-disaster communication scenario in which multiple UAV swarms relay data from a ground device in the postdisaster area to a remote access point (AP). The objective is to maximize the transmission rate
of the communication network through traffic routing design and controlling the excitation current weights and placements of UAVs. The optimal traffic routing is then derived using Ford-Fulkerson algorithm. Based on this routing, a DM-based particle swarm optimization algorithm (DM-PSO) is proposed to solve a variant of the optimization problem. Numerical experiments show that DM-PSO outperforms several other optimization algorithms due to the introduction of the DM, which enables the more crowded agents
to escape local optima and explore a broader search space.






\subsection{URLLC}
Given its effectiveness of access control, DM can be used for optimizing the blocklength of sensors in an URLLC-enabled sensor network as presented in~\cite{darabi2024diffusion}. The objective is to minimize the total energy consumption over the sensors. A dataset is first collected by solving the problem based on a traditional optimization algorithm with different fading channels, which is used for training the DM in which the CSI is used as conditional information, while the corresponding optimal blocklength value is used as the input state. The reverse process aims to predict the noise to minimize the difference between the actual and predicted noise. Simulation results show that with $60$ sensors, the proposed DM reduces the total power consumption up to $7$\% compared with the branching DQN and dueling double DQN. 


Different from \cite{darabi2024diffusion}, the work in \cite{wang2025effective} leverages the NOMA for the sensor access, whose problem is to jointly optimize the blocklength, power allocation, and decoding error probability to maximize the total throughput. A convex approximation scheme is developed to optimize the blocklenth, and DM-based DRL is utilized to optimize the power allocation and decoding error probability. The environment for the DRL agent encompasses the channel condition, number of sensors, maximum transmit power, and decoding reliability matrix. Simulation results show that the proposed DM-DRL can improve the effective throughput up to $38$\% and $42$\% compared with the standard DQN and DM-DRL with average blocklength allocation, respectively. 

\section{Conclusions, Existing Technical Issues, and Future Works} \label{sec:conclusions}
This paper has presented a comprehensive survey of the applications of DMs for future networks and communication systems. Particularly, we have provided detailed reviews, discussions, comparisons and important insights into the DM-based methods for emerging issues for future networks and communication systems. These include channel modeling and estimation, signal detection and data reconstruction, ISAC systems, resource management for edge computing networks, semantic communications and other emerging issues. The existing approaches show that DMs will be effective solutions to solve complicated problems in the future networks and communication systems. However, DMs still face technical limitations, which need to be further investigated in future works. 
\begin{itemize}
 \item \textit{High computational complexity and latency:} DMs typically require a large number of denoising steps to generate high-quality outputs, which leads to significant computational overhead and latency. The DMs have some difficulty in meeting tight constraints mission-critical services like ultra reliable low latency communications (URLLC) or autonomous driving due to the lack of real-time inference. Recent works have proposed new DM techniques, e.g., DiffuserLite~\cite{dong2024diffuserlite}, that shows some potential of realizing such real-time inference. 
    \item \textit{Real-world dataset in wireless networks:} Most existing works on DMs for wireless communications are trained and validated using simulated datasets, which can limit the practical effectiveness of DMs when deployed in real wireless systems. Thus, future researches should focus on building large-scale real-world datasets for training and benchmarking DM-based methods in realistic scenarios, which captures complex wireless dynamics such as fading channels, user mobility and unpredictable interference.
   
    \item \textit{Edge general intelligence:} Current research primarily employs DMs in task-specific domains within wireless communications, leaving substantial space for investigating DMs in edge general intelligence. Future research directions include developing DM-based architectures capable of all-purpose decision-making at edge nodes, enabling adaptive resource management, data reconstruction and inference across dynamically evolving network conditions. To achieve this, novel methodologies should focus on lightweight DMs maintaining high performance with reduced computational complexity, along with efficient fine-tuning mechanisms facilitating rapid adaptation to varying local network states \cite{10879580}.
    \item \textit{Customized user-intent networking:} Although existing DM applications broadly target system-wide optimization scenarios in wireless networks, future efforts should be put in DM-based frameworks tailored explicitly to individual user intents and preferences. Such customized user-intent-driven networking would require the development of novel DM-based inference techniques that dynamically learn and predict user-specific needs, enabling proactive and personalized network optimizations in terms of resource allocation, data prioritization, and interference management \cite{11049053}. Other critical challenges include balancing personalized model accuracy and computational efficiency, as well as ensuring real-time adaptability and user data privacy.  
\end{itemize}

\begin{IEEEbiography}[{\includegraphics[width=1.3in,height=1.3in,clip,keepaspectratio]{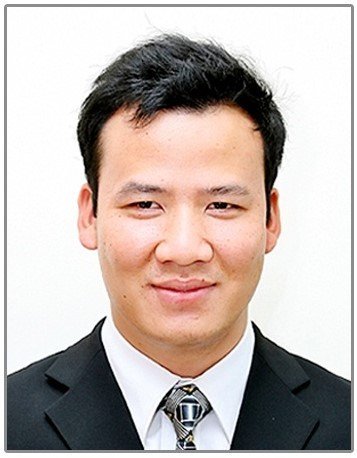}}] {Nguyen Cong Luong} received the B.S. and M.S. degrees from the School of Electrical and Electronic Engineering, Hanoi University of Science and Technology (HUST), Vietnam, and the Ph.D. degree from Institut Galilée, Université Sorbonne Paris Nord, France. His research interests include resource allocation and security management in the next generation networks.
\end{IEEEbiography}

\begin{IEEEbiography}[{\includegraphics[width=1in,height=1.25in,clip,keepaspectratio]{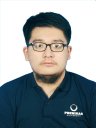}}] {Nguyen Duc Hai} (Student Member, IEEE) is an undergraduate student from the Phenikaa School of Computing, Phenikaa University, Vietnam. His research interest includes deep learning, deep reinforcement learning, game theory and incentive mechanisms for computer networks and communication networks.
\end{IEEEbiography}

\begin{IEEEbiography}[{\includegraphics[width=1in,height=1.25in,clip,keepaspectratio]{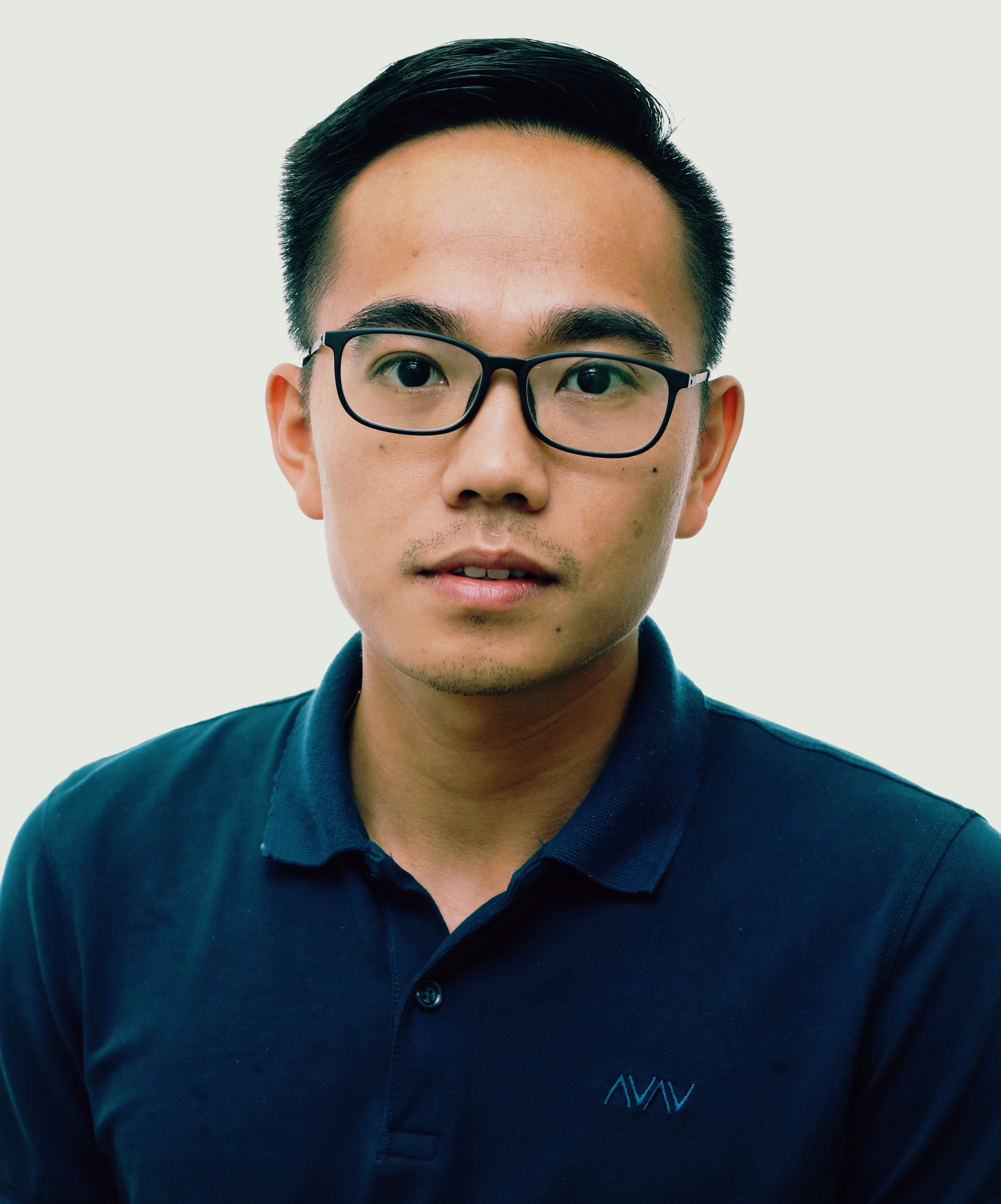}}]{Duc Van Le} (Senior Member, IEEE) received the BEng (Distinction) degree in electronics and telecommunications engineering from Le Quy Don Technical University, Vietnam, in 2011, and the PhD degree in computer engineering from University of Ulsan, South Korea, in 2016. Currently, he is a Postdoctoral Fellow with School of Electrical Engineering and Telecommunications, University of New South Wales (UNSW), Australia. During 2016-2024, he was a Senior Research Fellow with Nanyang Technological University (NTU), and a Research Fellow with National University of Singapore (NUS). His research interests include sensor networks, internet of things, cyber-physical systems, and applied machine learning. 
He was a recipient of the ACM/IEEE ICCPS'23 Best Paper Award.
\end{IEEEbiography}

\begin{IEEEbiography}[{\includegraphics[width=1in,height=1.25in,clip,keepaspectratio]{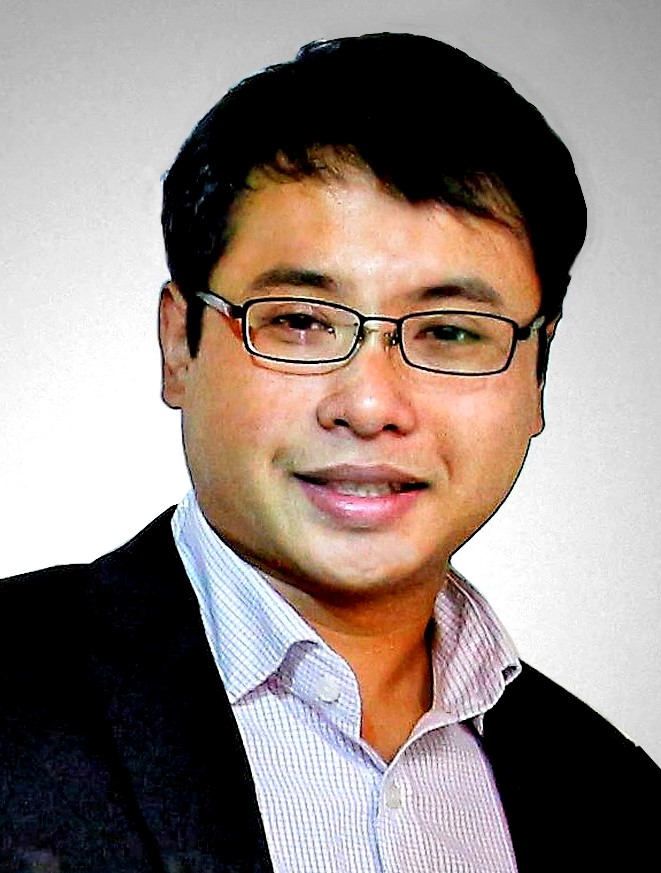}}]{Huy T. Nguyen} (Member, IEEE) received his B.Eng. (Hons.) in Computer Science and Engineering from HCMC University of Technology, Vietnam, in 2013. He then earned his M.Eng. and Ph.D. degrees from the Department of Information and Communication System, Inje University, South Korea, in 2016 and 2019, respectively.  He is currently a Lecturer at the Faculty of Information Technology, Van Lang University (VLU), where he also leads the Smart and Autonomous Systems (SAS) Research Group. Before joining VLU, he served as Chief Technology Officer (CTO-Asia Desk) at Data Design Engineering (DDE), primarily developing data-driven AI solutions. His earlier career included significant contributions to the communication aspects of wireless SmartGrid systems as a Scientist at the Institute for Infocomm Research (I2R-A*STAR) and Nanyang Technological University (NTU), Singapore. His research interests lie in applying AI/ML and optimization techniques to signal processing and wireless communications for 5G and beyond.
\end{IEEEbiography}

\begin{IEEEbiography}
[{\includegraphics[width=1in,height=1.25in,clip,keepaspectratio]{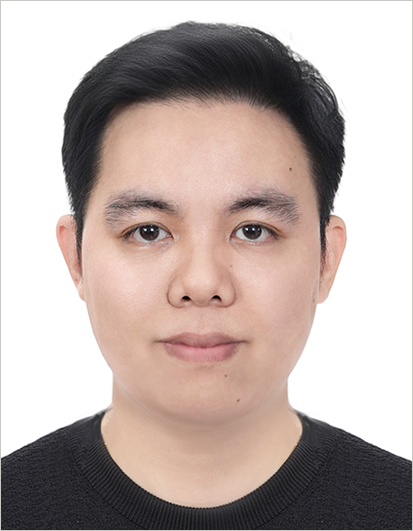}}]
 {Thai-Hoc Vu} (Member, IEEE) is currently a Lecturer at the Institute of Information Technology, Digital Transformation, Thu Dau Mot University, Binh Duong 820000, Vietnam. He earned his Ph.D. degree in Electrical Engineering from the University of Ulsan, Korea, in 2025.

He specialises in the performance analysis of wireless communications systems, focusing on cutting-edge areas such as non-orthogonal multiple access, reconfigurable intelligent surface, radio-frequency-based energy harvesting, physical-layer security, short-packet transmission, and applying convex optimisation and machine learning to analyse, improve, and optimise advanced wireless communication networks and IoT systems. He has authored and co-authored more than 50 IEEE transactions/Journals and flagship IEEE conference papers. He was awarded the Vietnamese Young Scientists in Korea Award from the Vietnamese Students Association in Korea and the Vietnam Embassy in the Republic of Korea in 2022. He completed his Ph.D. program with more than 20 Top IEEE publications as the first author and received the best research award from BK21 five times from 2021 to 2025. He was a recipient of the Top Reviewer Award from IEEE Communications Letters in 2023 and 2024, IEEE ICCE Best Paper Student Awards in 2022 and 2024, and IEEE ATC Best Paper Award in 2024.
\end{IEEEbiography}

\begin{IEEEbiography}[{\includegraphics[width=1in,height=1.25in,clip,keepaspectratio]{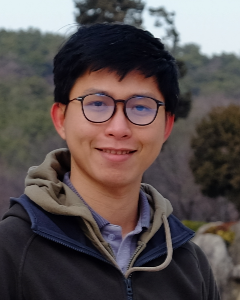}}]{Thien Huynh-The} (Senior Member, IEEE) received the Ph.D. degree in Computer Science and Engineering from Kyung Hee University (KHU), South Korea, in 2018. He was a recipient of the Superior Thesis Prize awarded by KHU. From March 2018 to August 2018, he was a Postdoctoral Researcher with Ubiquitous Computing Laboratory, KHU. From September 2018 to May 2022, he was a Postdoctoral Researcher with ICT Convergence Research Center, Kumoh National Institute of Technology, South Korea. He is currently a Lecturer in Department of Computer and Communications Engineering, Ho Chi Minh  City University of Technology and Education, Vietnam. He was a recipient of Golden Globe Award 2020 for Vietnamese Young Scientist and IEEE ATC Best Paper Award in 2023. His current research interests include digital image processing, radio signal processing, computer vision, wireless communications, IoT applications, machine learning, and deep learning. He is currently serving as an Editor of IEEE COMML.
\end{IEEEbiography}

\begin{IEEEbiography}[{\includegraphics[width=1in,height=1.25in,clip,keepaspectratio]{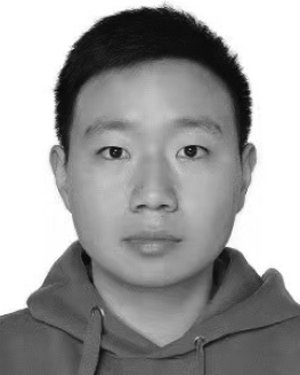}}]{Ruichen Zhang} (Member, IEEE) received the B.E. degree from Henan University, China, in 2018, and the Ph.D. degree from Beijing Jiaotong University, China, in 2023. He is currently a Postdoctoral Research Fellow with the College of Computing and Data Science, Nanyang Technological University, Singapore. In 2024, he was a Visiting Scholar with the College of Information and Communication Engineering, Sungkyunkwan University, Suwon, South Korea. His research interests include LLM-empowered networking, reinforcement learning-enabled wireless communication, generative AI models, and heterogeneous networks. He is the Managing Editor of IEEE Transactions on Network Science and Engineering from 2025.
\end{IEEEbiography}

\begin{IEEEbiography}[{\includegraphics[width=1in,height=1.25in,clip,keepaspectratio]{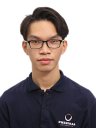}}] {Nguyen Duc Duy Anh} is an undergraduate student from the Phenikaa School of Computing, Phenikaa University, Vietnam. His research interest includes deep learning, deep reinforcement learning, game theory and incentive mechanisms for computer networks and communication networks.
\end{IEEEbiography}

\begin{IEEEbiography}[{\includegraphics[width=1.2in,height=1.2in,clip,keepaspectratio]{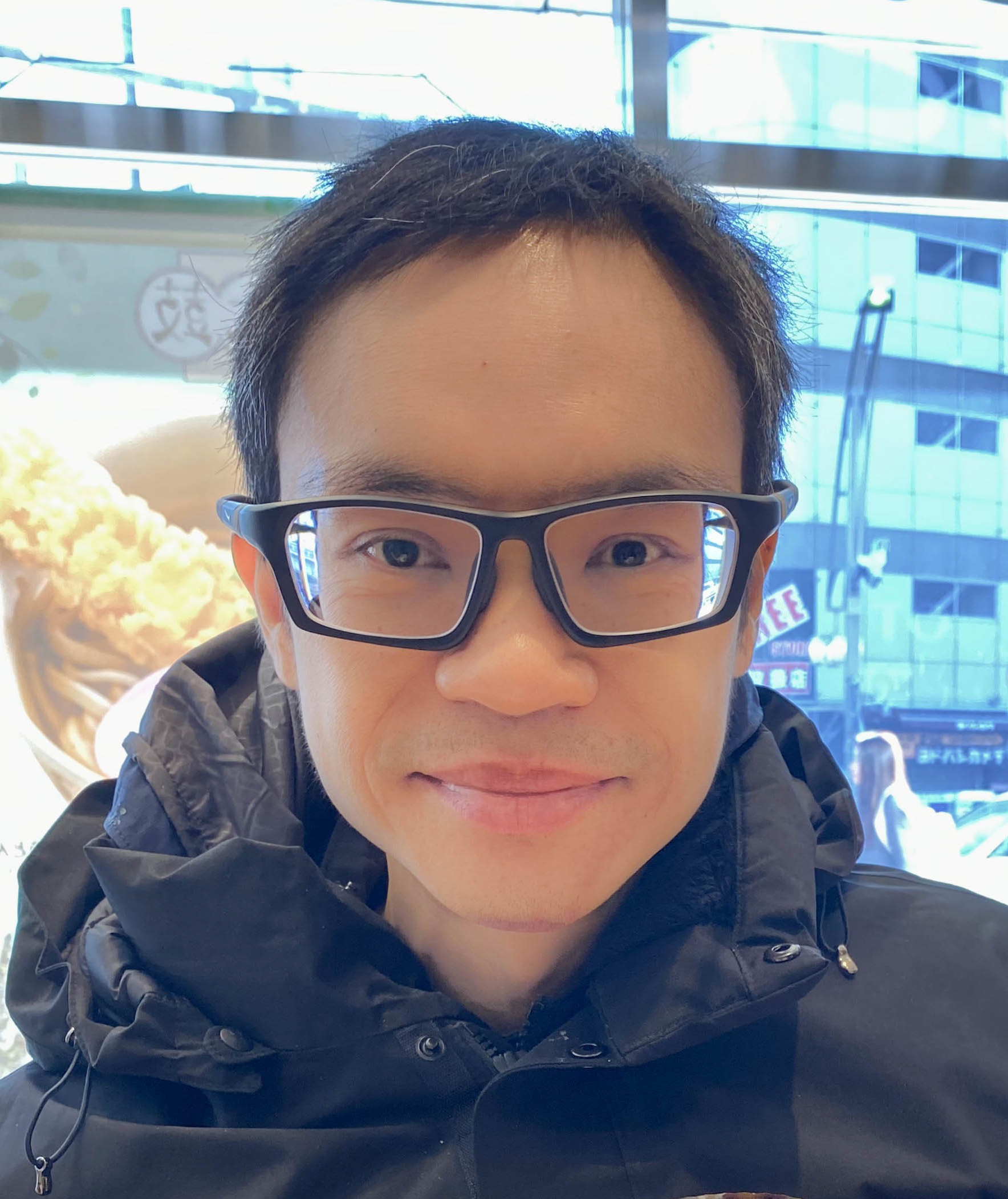}}] {Dusit Niyato} (Fellow, IEEE) received the B.Eng. degree from the King Mongkut's Institute of Technology Ladkrabang, Bangkok, Thailand, in 1999 and the Ph.D. degree in electrical and computer engineering from the University of Manitoba, Winnipeg, MB, Canada, in 2008. He is currently a Professor with the School of Computer Science and Engineering, Nanyang Technological University, Singapore. His research interests include energy harvesting for wireless communication, Internet of Things (IoT), and sensor networks.
\end{IEEEbiography}

\begin{IEEEbiography}[{\includegraphics[width=1in,height=1.25in,clip,keepaspectratio]{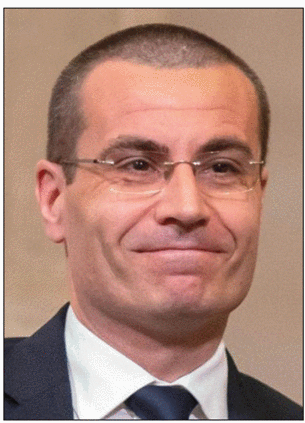}}]{Marco Di Renzo} [F] is a CNRS Research Director (Professor) and the Head of the Intelligent Physical Communications group with the Laboratory of Signals and Systems at CNRS \& CentraleSupelec, Paris-Saclay University, Paris, France, as well as a Chair Professor in Telecommunications Engineering with the Centre for Telecommunications Research, Department of Engineering, King's College London, London, United Kingdom.

He served as the Editor-in-Chief of \textit{IEEE Communications Letters} from 2019 to 2023. His current main roles within the IEEE Communications Society include serving as a Voting Member of the Fellow Evaluation Standing Committee, as the Chair of the Publications Misconduct Ad Hoc Committee, and as the Director of Journals. He is also on the Editorial Board of the \textit{Proceedings of the IEEE}.
\end{IEEEbiography}

\begin{IEEEbiography}[{\includegraphics[width=1.1 in,height=1.3 in,clip,keepaspectratio]{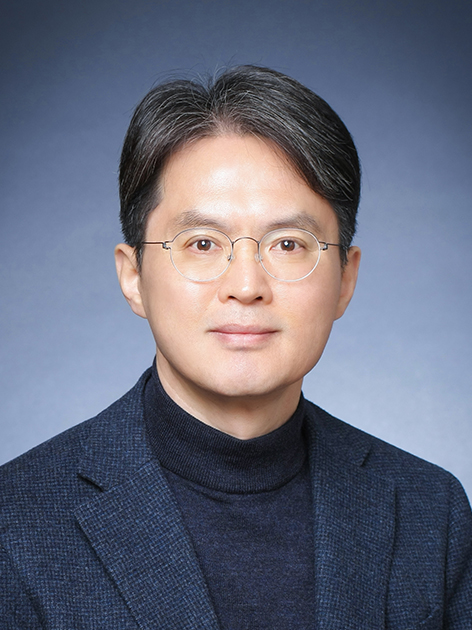}}] {Dong In Kim} (Fellow, IEEE) received the Ph.D. degree in electrical engineering from the University of Southern California, Los Angeles, CA, USA, in 1990. He was a Tenured Professor with the School of Engineering Science, Simon Fraser University, Burnaby, BC, Canada. He is a Distinguished Professor with the College of Information and Communication Engineering, Sungkyunkwan University, Suwon, South Korea. He is a Fellow of the Korean Academy of Science and Technology and a Member of the National Academy of Engineering of Korea. He was a first recipient of the NRF of Korea Engineering Research Center in Wireless Communications for RF Energy Harvesting, from 2014 to 2021. He has been listed as a 2020/2022 Highly Cited Researcher by Clarivate Analytics. Since 2001, he has been serving as an editor, editor at large, and Area Editor of Wireless Communications I for the IEEE Transactions on Communications. From 2002 to 2011, he also served as an editor and a Founding Area Editor of Cross-Layer Design and Optimization for the IEEE Transactions on Wireless Communications. From 2008 to 2011, he served as the Co-Editor-in-Chief for the IEEE/KICS Journal of Communications and Networks. He served as the Founding Editor-in-Chief for the IEEE Wireless Communications Letters, from 2012 to 2015. He was selected the 2019 recipient of the IEEE Communications Society Joseph LoCicero Award for Exemplary Service to Publications. He was the General Chair for IEEE ICC 2022 in Seoul.
\end{IEEEbiography}	

\begin{IEEEbiography}[{\includegraphics[width=1in,height=1.25in,clip,keepaspectratio]{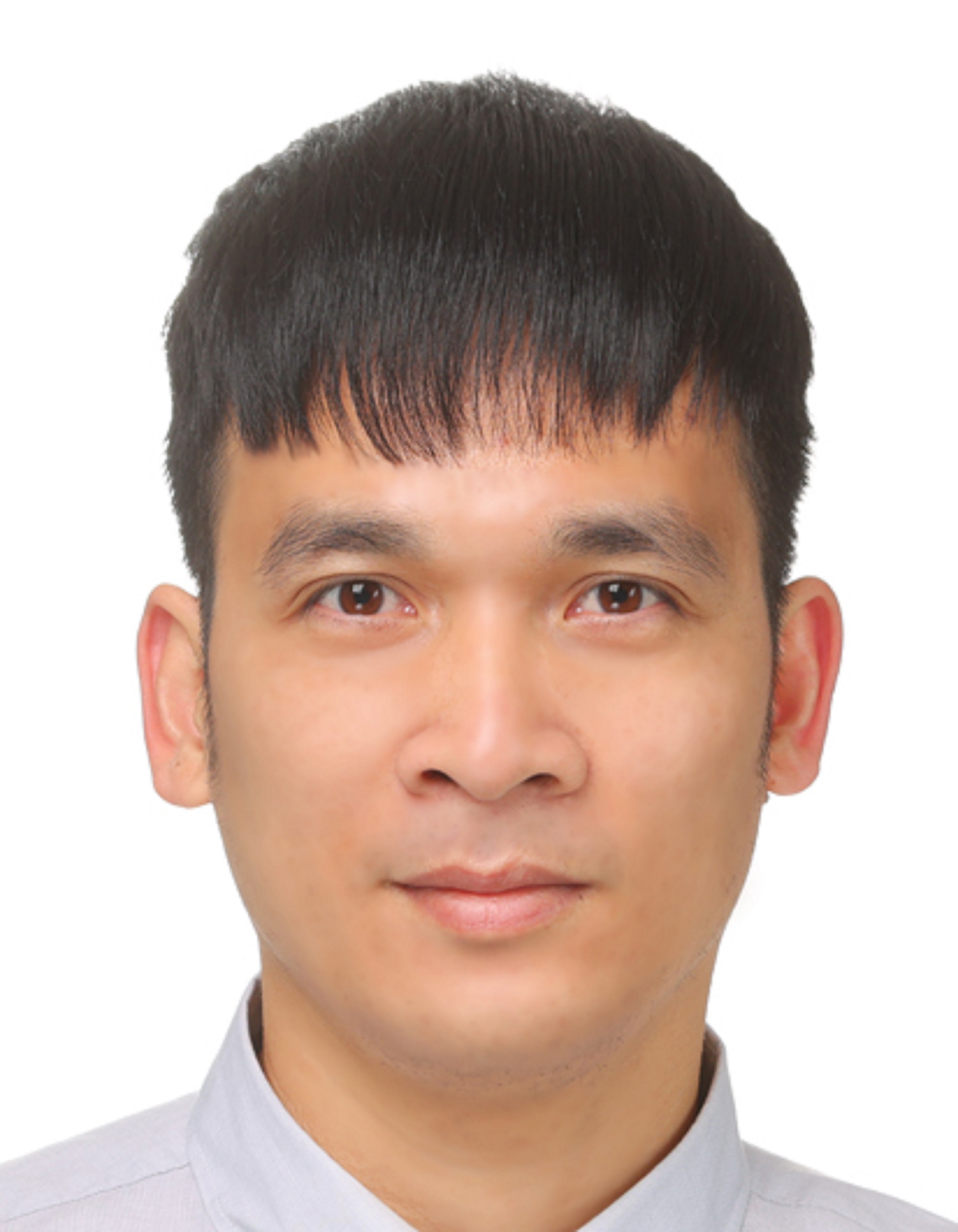}}]{Quoc-Viet Pham} (Senior Member, IEEE) is currently an Assistant Professor in Networks and Distributed Systems at the School of Computer Science and Statistics, Trinity College Dublin, Ireland. He earned his BSc and PhD degrees (with Best PhD Dissertation Award) in Telecommunications Engineering from Hanoi University of Science and Technology and Inje University in 2013 and 2017, respectively.

He has special research interests in the areas of wireless AI, edge computing, Internet of Things, and distributed learning. He was a recipient of the IEEE TVT Top Reviewer Award in 2020, Golden Globe Award in Science and Technology for Vietnam's Young Researchers in 2021, IEEE ATC Best Paper Award in 2022, and IEEE MCE Best Paper Award in 2023, IEEE M-COMSTD Exemplary Editor Award in 2024, and Clarivate Highly Cited Researcher Award in 2024. He was honoured with the IEEE ComSoc Best Young Researcher Award for EMEA 2023 in recognition of his research activities for the benefit of the Society.

He was the Lead Guest Editor of the IEEE Internet of Things Journal special issue on Aerial Computing for the Internet of Things. He currently serves as an Editor of IEEE Communications Letters, IEEE Communications Standards Magazine, IEEE Communications Surveys \& Tutorials, Journal of Network and Computer Applications, IEEE Transactions on Mobile Computing, and IEEE Transactions on Network Science and Engineering.
\end{IEEEbiography}

\end{document}